\documentclass{article}

\usepackage{microtype}
\usepackage{graphicx}
\usepackage{subfigure}
\usepackage{booktabs} 

\usepackage[pagebackref,breaklinks,colorlinks,citecolor=cvprblue]{hyperref}

\usepackage[accepted]{icml2024}

\usepackage{amsmath}
\usepackage{amssymb}
\usepackage{mathtools}
\usepackage{amsthm}

\usepackage[capitalize,noabbrev]{cleveref}
\crefname{definition}{Def.}{Defs.}
\crefname{namedtheorem}{Theorem}{Theorems}

\usepackage[acronym]{glossaries}
\usepackage{graphicx}
\usepackage{wrapfig}
\usepackage{tabularx}
\usepackage{booktabs} 
\usepackage{pifont} 
\usepackage{multirow} 
\usepackage{appendix}

\definecolor{Gain}{RGB}{34,139,34}
\definecolor{Drop}{RGB}{189,30,9}

\theoremstyle{plain}
\newtheorem{theorem}{Theorem}[section]

\newtheorem{lemma}[theorem]{Lemma}

\theoremstyle{definition}
\newtheorem{definition}[theorem]{Definition}

\theoremstyle{remark}

\usepackage[textsize=tiny]{todonotes}

\usepackage[subtle, 
]{savetrees}

\graphicspath{{./figures/}}

\newglossary[slg]{symboltype}{syi}{syg}{List of Symbols}

\newglossaryentry{ebm}{
    type=\acronymtype, name={EBM},
    description={Energy-Based Model},
    longplural={Energy-Based Models},
    first={Energy-Based Models (EBM)}, 
    firstplural={Energy-Based Models (EBMs)}
}
\newglossaryentry{its}{
    type=\acronymtype, name={ITS},
    description={Inverse Transformation Search},
    longplural={Inverse Transformation Searches},
    first={Inverse Transformation Search (ITS)}, 
    firstplural={Inverse Transformation Searches (ITSs)}
}
\newglossaryentry{stn}{
    type=\acronymtype, name={STN},
    description={Spatial Transformer Network},
    longplural={Spatial Transformer Networks},
    first={Spatial Transformer Network (STN)}, 
    firstplural={Spatial Transformer Networks (STNs)}
}
\newglossaryentry{dtn}{
    type=\acronymtype, name={DTN},
    description={Diffeomorphic Transformer Network},
    longplural={Diffeomorphic Transformer Networks},
    first={Diffeomorphic Transformer Network (DTN)}, 
    firstplural={Diffeomorphic Transformer Networks (DTNs)}
}
\newglossaryentry{cnn}{
    type=\acronymtype, name={CNN},
    description={Convolutional Neural Network},
    longplural={Convolutional Neural Networks},
    first={Convolutional Neural Network (CNN)}, 
    firstplural={Convolutional Neural Networks (CNNs)}
}
\newglossaryentry{gcnn}{
    type=\acronymtype, name={GCNN},
    description={Group CNN},
    longplural={Group CNNs},
    first={Group CNN (GCNN)}, 
    firstplural={CGroup CNNs (GCNNs)}
}
\newglossaryentry{fcn}{
    type=\acronymtype, name={FCN},
    description={Fully-Connected Network},
    longplural={Fully-Connected Networks},
    first={Fully-Connected Network (FCN)}, 
    firstplural={Fully-Connected Networks (FCNs)}
}
\newglossaryentry{gge}{
    type=\acronymtype, name={GGE},
    description={Group-Generator Encoder},
    longplural={Group-Generator Encoders},
    first={Group-Generator Encoder (GGE)}, 
    firstplural={Group-Generator Encoders (GGEs)}
}
\newglossaryentry{vit}{
    type=\acronymtype, name={ViT},
    description={Vision Transformer},
    longplural={Vision Transformers},
    first={Vision Transformer (ViT)}, 
    firstplural={Vision Transformers (ViTs)}
}
\newglossaryentry{etn}{
    type=\acronymtype, name={ETN},
    description={Equivariant Transformer Network},
    longplural={Equivariant Transformer Networks},
    first={Equivariant Transformer Network (ETN)}, 
    firstplural={Equivariant Transformer Networks (ETNs)}
}

\newglossaryentry{symbol:image}{
    name=\ensuremath{\boldsymbol{X}},
    description={Image},
    type=symboltype
}
\newglossaryentry{symbol:image_rec}{
    name=\ensuremath{\boldsymbol{X}_{\text{rec}}},
    description={Reconstructed Image},
    type=symboltype
}
\newglossaryentry{symbol:image_orbit}{
    name=\ensuremath{\boldsymbol{X}_{\text{orb}}},
    description={Orbit Stack},
    type=symboltype
}
\newglossaryentry{symbol:latent_dim}{
    name=\ensuremath{L},
    description={Latent},
    type=symboltype
}
\newglossaryentry{symbol:channel_dim}{
    name=\ensuremath{C},
    description={Channels},
    type=symboltype
}
\newglossaryentry{symbol:height_dim}{
    name=\ensuremath{H},
    description={Height},
    type=symboltype
}
\newglossaryentry{symbol:width_dim}{
    name=\ensuremath{W},
    description={Width},
    type=symboltype
}
\newglossaryentry{symbol:dataset}{
    name=\ensuremath{\mathcal{X}},
    description={Dataset},
    type=symboltype
}
\newglossaryentry{symbol:dataset_train}{
    name=\ensuremath{\mathcal{X}_{\text{train}}},
    description={Trainingset},
    type=symboltype
}
\newglossaryentry{symbol:dataset_test}{
    name=\ensuremath{\mathcal{X}_{\text{test}}},
    description={Testset},
    type=symboltype
}
\newglossaryentry{symbol:encoder}{
    name=\ensuremath{f}, 
    description={Encoder},
    type=symboltype
}
\newglossaryentry{symbol:decoder}{
    name=\ensuremath{p}, 
    description={Decoder},
    type=symboltype
}
\newglossaryentry{symbol:gge}{
    name=\ensuremath{g_{\boldsymbol{\phi}}},
    description={Encoder},
    type=symboltype
}
\newglossaryentry{symbol:emb}{
    name=\ensuremath{\boldsymbol{z}},
    description={Embedding},
    type=symboltype
}
\newglossaryentry{symbol:weights}{
    name=\ensuremath{\boldsymbol{W}},
    description={Weight Matrix},
    type=symboltype
}
\newglossaryentry{symbol:loss}{
    name=\ensuremath{\mathcal{L}},
    description={Loss},
    type=symboltype
}
\newglossaryentry{symbol:transparam}{
    name=\ensuremath{\boldsymbol{\theta}},
    description={Transformation},
    type=symboltype
}
\newglossaryentry{symbol:transform}{
    name=\ensuremath{\boldsymbol{T}},
    description={Transformation},
    type=symboltype
}
\newglossaryentry{symbol:all_transforms}{
    name=\ensuremath{\mathcal{T}},
    description={All Transformations considered},
    type=symboltype
}
\newglossaryentry{symbol:transform_pred}{
    name=\ensuremath{\hat{\gls*{symbol:transform}}},
    description={Predicted Transformation},
    type=symboltype
}
\newglossaryentry{symbol:transfunc}{
    name=\ensuremath{\zeta_{\gls*{symbol:transparam}}},
    description={Transformation Function},
    type=symboltype
}
\glsunsetall  

\icmltitlerunning{Tilt your Head: Activating the Hidden Spatial-Invariance of Classifiers}

\begin{document}

\twocolumn[
\icmltitle{Tilt your Head: Activating the Hidden Spatial-Invariance of Classifiers}



\icmlsetsymbol{equal}{*}

\begin{icmlauthorlist}
\icmlauthor{Johann Schmidt}{yyy}
\icmlauthor{Sebastian Stober}{yyy}
\end{icmlauthorlist}

\icmlaffiliation{yyy}{Artificial Intelligence Lab, Otto-von-Guericke University, Magdeburg, Germany}

\icmlcorrespondingauthor{Johann Schmidt}{johann.schmidt@ovgu.de}

\icmlkeywords{Canonicalisation, Geometric Deep Learning, Computer Vision}

\vskip 0.3in
]



\printAffiliationsAndNotice{} 

\begin{abstract}
Deep neural networks are applied in more and more areas of everyday life.
However, they still lack essential abilities, such as robustly dealing with spatially transformed input signals.
Approaches to mitigate this severe robustness issue are limited to two pathways:
Either models are implicitly regularised by increased sample variability (data augmentation) or explicitly constrained by hard-coded inductive biases.
The limiting factor of the former is the size of the data space, which renders sufficient sample coverage intractable.
The latter is limited by the engineering effort required to develop such inductive biases for every possible scenario.
Instead, we take inspiration from human behaviour, where percepts are modified by mental or physical actions during inference.
We propose a novel technique to emulate such an inference process.
This is achieved by traversing a sparsified inverse transformation tree during inference using parallel energy-based evaluations.
Our proposed inference algorithm, called Inverse Transformation Search (ITS), is model-agnostic and equips the model with zero-shot pseudo-invariance to spatially transformed inputs.
We evaluated our method on several benchmark datasets, including a synthesised ImageNet test set.
ITS outperforms the utilised baselines on all zero-shot test scenarios.
\end{abstract}    
\section{Introduction}
\label{sec:intro}

A primary objective in Deep learning is to develop models that exhibit robust generalisation during inference (commonly evaluated using test sets). To improve generalisation, models are trained with data sets of increasing variability and size. However, this often requires using larger models \cite{Chefer2022}. As novel datasets more closely mirror real-world scenarios, models undergo rigorous assessments in diverse, dynamic environments.
This requires the models to learn intricate functions that generalise to all the variations presented in the data.
It was shown that modern vision models break even under minor input perturbations \cite{Engstrom2019}, posing severe risks for real-world applications.
Each new variation (dimension) scales the data space exponentially, known as the \emph{combinatorial explosion}. Sampling in such spaces gets progressively harder (\emph{curse of dimensionality}). 
For instance, consider an object from infinitely many perspectives.
In a standard image classification task, all these variations have the same label assigned.
Using 2D spatial transformations, these variations bring only redundant information.
Hence, considering every conceivable variant is impractical, computationally infeasible, and even useless in terms of information gain.
If a model can correctly predict an object from all perspectives, it is \emph{(pseudo-)invariant} to these perspective changes.





\begin{figure}[t]
\begin{center}
\centerline{\includegraphics[width=\columnwidth]{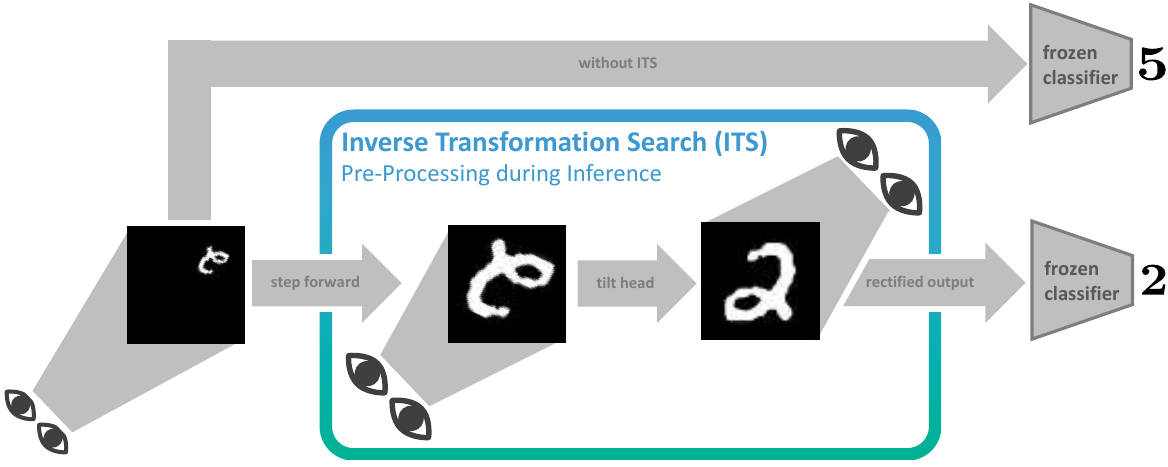}}
\caption{Inspired by the human recognition process for perturbed inputs, we propose a novel inference framework coined \glsfirst*{its}. 
Similar to humans transforming the given input mentally or physically by tilting their heads or stepping forward, for instance.
Instead of the usual likelihood estimation of a given sample, the pre-trained model can evaluate different input transformations before a final decision is made.
The output of this process is the most familiar transformation of the query.}
\vspace{-1cm}
\label{fig:summary}
\end{center}
\end{figure}

\paragraph{Data-driven Pseudo-Invariance}
Over recent years, data augmentation \cite{Hernandez2018, Shorten2019} has become a routine instrument for increasing generalisation.
The main drivers for this development are its simplicity and model agnosticism.
Data augmentation enriches a dataset by applying transformations or modifications to existing data points.
Augmented training data acts as an explicit Frobenius norm regularizer \cite{LeJeune2019} promoting the extraction of consistent class features \cite{Huang2021}.
Models either learn transformed replicates of the same filter \cite{olah2020} or fuzzy averages of transformed per-class features, regularising model complexity \cite{Chen2019InvarianceRV, kernelDA2019}.
This behaviour can be amplified using invariance-promoting losses \cite{Shakerinava2022}. 

\citet{Goodfellow2009} showed that the quality of pseudo-invariance in the learned representation increases with the depth of the network. Minor deformations, like small shifts and rotations, are compensated in lower layers, while higher layers progressively form invariances.
Augmented training can encourage this behaviour, as the model aims to align the kernel space with the largest eigenvector of the sample covariance \cite{Balestriero2022ADI}.
However, strong forms of data augmentation, like cropping, can also hurt generalization as it can introduce feature biases by varying the dominance of features \cite{Balestriero2022}.

\paragraph{Model-driven (Pseudo-)Invariance}
Inductive biases for spatial symmetries are commonly developed for \glspl*{cnn} by exploiting their weight-shared filters \cite{Schmidt2022}.
Instead of regularising the filter space using data, it can be constrained explicitly to symmetric weight matrices (so-called isotropic filters) \cite{hu2019exploring}.
Invariance can also be achieved without such constraints by considering all transformed filter variations followed by a max-out \cite{Gens2014, cohen2016group, bao2020}.
This can be formulated more flexibly by a steerable basis for the space of convolutional kernels \cite{Weiler2017}.
This allows for weight sharing across filter orientation without aliasing artifacts.
Recently, various options for such basis functions were proposed, most of which are derived from circular symmetries.
This includes Bessel functions \cite{dcfnet, delchevalerie2021}, Hermite polynomials \cite{ecker2018rotationequivariant, Ustyuzhaninov2020Rotation} and irreducible representations to steer the harmonics \cite{Weiler2019}.
\citet{Zhou2020} proposed to meta-learn the filter structure (binary mask) based on the equivariance required to solve a task.
Recently, it was shown that equivariant models can be \emph{canonicalised} by fitting a prior over the group during training \cite{Kaba2023, Mondal2023}.

Unfortunately, these inductive biases come with a significant engineering effort, which renders a generalisation to other symmetries cumbersome.
The same applies to methods, which learn the geometry of the latent space \cite{Arvanitidis2017} to find equivariance paths.
With the introduction of \glsfirst*{stn} \cite{jaderberg2016spatial}, another promising avenue began.
These networks remove the spatial perturbations, i.e., \emph{canonicalise} the transformed input signals by learning a highly non-linear mapping from pixel to transformation space.
A more effective alternative is the mapping to canonical coordinate spaces \cite{PolarTransformer, tai2019equivariant}.
To learn transformation compositions, \glspl*{stn} can be stacked \cite{Lin2017, Detlefsen2018}.
These transformations can also act on each pixel individually using pixel-wise motion fields \cite{Shu2018}.
However, all of these variants require augmented training data.
\textit{We propose a novel inference strategy that canonicalises inputs without augmented training.}

\paragraph{Inference-driven Pseudo-Invariance}
We take inspiration from human behaviour and propose a third way to integrate pseudo-invariances.
The human brain learns to discard redundancies imposed by symmetries for a more efficient perception \cite{Gomez2008, Leibo2011}.
To recognise a perceived object, our brain aligns it with our learned mental model of that object.
This process is called \emph{canonicalisation} \cite{Graf2006} and it is either performed mentally or physically (or a mixture of both).
We can leverage our physical abilities by squeezing our eyes, tilting our heads and slowly moving closer when triggered by an unclear perception.
In this way, our trained mental model can properly recognise a variety of perturbed inputs by gradually aligning them to their known canonical form.
This is a well-established theory supported by various experiments \cite{palmer1981, Harris2001, Gomez2008, Konkle2011}.

We can transfer this hypothesis by emulating the alignment search algorithmically as illustrated in \cref{fig:summary}.
\emph{We propose an \glsfirst*{its} for trained classifiers during inference, which renders them zero-shot pseudo-invariant by only requiring access to their logit scores.}
We disentangle the process of class-specific feature learning (achieved during training) and canonical signal distillation (achieved during inference).
The latter removes perturbations from the input and therefore eliminates variances from the query to ease the subsequent prediction.

\paragraph{Scope and Outline}
In this work, we particularly focus on 2D image classifiers and the special linear group, comprising combinations of rotation, scaling, and shearing operations.
However, our proposed method is more generally applicable.
In \cref{sec:preliminaries}, we start with an introduction to the language of symmetries, i.e., group theory.
Upon these fundamentals, we build a group-induced evaluation score in \cref{sec:confidence} used in our search algorithm presented in \cref{sec:methodology}.
We evaluate our methods in \cref{sec:experiments} and conclude with future directions in \cref{sec:conclusion}.

\section{Preliminaries}
\label{sec:preliminaries}

\begin{figure}
    \centering
    \includegraphics[width=0.45\textwidth]{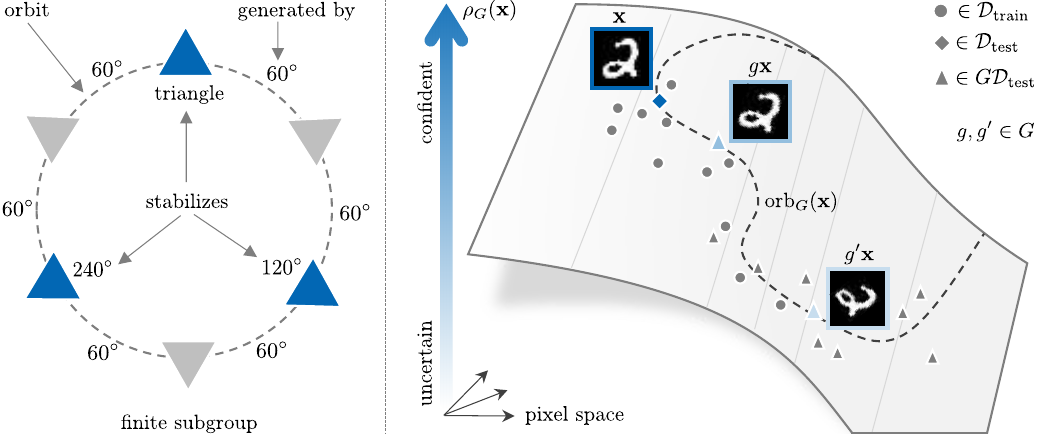}
    \caption{\emph{(left)} Centre-Rotating an equilateral triangle and \emph{(right)} an exemplary confidence surface.}
    \label{fig:manifold}
\end{figure}

Understanding how to undo transformations requires a basic comprehension of symmetries and their mathematical formulations as groups.
A \emph{symmetry} of an object is a transformation that preserves the object's structure.
For instance, rotating an equilateral triangle around its centre by $0^\circ$, $120^\circ$ or $240^\circ$ leaves the triangle unaffected, as illustrated in \cref{fig:manifold} (left).
These rotations form a \emph{group} of finite size, i.e., a \emph{finite group}.
\begin{definition}[Group] \label{def:axioms}
Groups are sets of objects equipped with a binary operation.
For every group $G$ the following axioms must hold
\[
\begin{array}{ll}
    \text{closure:} & g,h \in G \rightarrow gh \in G, \\
    \text{associativity:} & g(hk) = (gh)k, \\
    \text{identity:} & eg = ge = g, \\
    \text{invertibility:} & gg^{-1} = g^{-1}g = e,
\end{array}
\] 
with $g,h,k,e \in G$ and $e$ being the identity element.
\end{definition}
This includes a composition of multiple group elements, like $gh$, which is the short form for $g \circ h$, where $\circ$ is some binary operator.
Some groups possess a fifth property, that is, \emph{commutativity}.
We call them \emph{Abelian groups}, e.g., centre-rotations. 
More details and proofs can be found in \cite{nash2016}.

Let us extend the picture from discrete to continuous structures. 
For a circle, centre rotations of any angle send the object back to itself.
Formally, these groups are called \emph{Matrix Lie groups} \cite{hall2015lie}, which form a smooth differentiable manifold.
For the basic comprehension of this work, it is only important to understand that group elements can be represented by parameterised matrices.
In this work, we focus on centre-rotation, isotropic-scaling, and shearing transformations (and their compositions).
All of them are represented by $2 \times 2$ matrices $\mathbf{T}$.
This particular group is called the Special Linear Group $\text{SL}_2(\mathbb{R})$ \cite{hall2015lie}.
\\
When dealing with continuous structures, it is helpful to consider finite approximations instead. 
Given a group $G$ we can have a \emph{subgroup} $G^\prime < G$, which still satisfies \cref{def:axioms} but with a smaller \emph{cardinality} $|G^\prime|$.
For instance, a subgroup of all centre-rotations is the finite group with multiples of $10^\circ$ rotations.
Such groups can be generated by a single group element.
\begin{definition}[Generator] \label{def:generator}
    Given $g \in G$, $g$ can be used as a generator to form a finite (cyclic) subgroup $\{ \langle g \rangle^k \mid k \in \mathbb{Z} \}$.
    $\langle g \rangle^k$ denotes $g$ to be a generator acting $k$-times. To remove visual clutter, we use $\hat{g}$ to denote the generating group element.
\end{definition}
A group element \emph{acts} on an object $\mathbf{x}$ by transforming it.
If the entire group $G$ acts on an object, the resulting output set is called the orbit of the object under $G$.
\begin{definition}[Orbit] \label{def:orbit}
    Let $G$ be a group and $\mathbf{x}$ an object. Then, the orbit of $\mathbf{x}$ under $G$ is
    \[
    \text{orb}_G(\mathbf{x}) = \{ g\mathbf{x} \mid g \in G \}.
    \]
\end{definition}
All $g$ that leave $\mathbf{x}$ unchanged are symmetries of that object.
\begin{definition}[Stabiliser] \label{def:stabilizer}
    The stabiliser of some $\mathbf{x}$ under $G$ is a subgroup that leaves $\mathbf{x}$ unchanged,
    \[
    \text{stab}_G(\mathbf{x}) = \{ g \in G \mid g\mathbf{x} = \mathbf{x}\}.
    \]
\end{definition}
We use stabilisers to formalise spatial symmetries, e.g., transformations that preserve the pixel positions of an image.
\\
We can also formalise this from a function perspective.
A function $f$ (e.g., modelled by a neural network), which captures a symmetry defined by group $G$, is either \emph{equivariant}, \emph{invariant}, or \emph{covariant} \cite{Marcos_2017}.
\begin{definition}[Variances] \label{def:variances}
    Let $f(\mathbf{x})$ be a function of an input $\mathbf{x}$.
    Then, $f$ is
    \[
    \begin{array}{llll}
        \text{equivariant} & \text{iff} & f(g \mathbf{x}) = g f(\mathbf{x}), & \forall g \in G, \\
        \text{invariant} & \text{iff} & f(g \mathbf{x}) = f(\mathbf{x}), & \forall g \in G, \\
        \text{covariant} & \text{iff} & f(g \mathbf{x}) = g^\prime f(\mathbf{x}), & \forall g \in G, g^\prime \in G^\prime,
    \end{array}
    \] 
    with $G^\prime$ being isomorphic to $G$.
\end{definition}
Note that covariance is the most general one, where both transformations differ in their domain and co-domain, respectively.
Invariance and equivariance are special cases of covariance.
If \cref{def:variances} does not hold but is approximately equal, we call the variance a \emph{pseudo-variance}.
For more elaborated details about variances in deep learning, see \cite{bronstein2021geometric}.

\section{Localizing Canonical Forms}
\label{sec:confidence}
In this section, we introduce a novel method for finding canonical forms.
We use $f_{\mathbf{\theta}}: \Omega \to \mathbb{R}^Y$ to denote a parameterised function mapping from pixel space to class-logits.
We call this function a softmax-classifier if it parameterises a categorical distribution by its logits, such that 
\begin{equation} \label{eq:boltzmann}
    p_{\mathbf{\theta}}(y \mid \mathbf{x}) = \operatorname{softmax}\left(f_{\mathbf{\theta}}(\mathbf{x})\right) := \frac{\exp \left(f_\theta(y \mid \mathbf{x})\right)}{\sum_{y^{\prime}} \exp \left(f_\theta(y^{\prime} \mid \mathbf{x})\right)}.
\end{equation}
During training, we obtain the posterior 
\begin{equation} \label{eq:posterior}
    p(y \mid \mathbf{x}, \mathcal{D}_{\text{train}}) = \int p_{\mathbf{\theta}}(y \mid \mathbf{x}) p(\mathbf{\theta} \mid \mathcal{D}_{\text{train}}) d \mathbf{\theta}.
\end{equation}

\paragraph{$G$-transformed Test Sets}
Commonly, models are tested on holdout sets with the same spatial variation.
Instead, we aim to train a classifier on canonical examples (vanilla training data) and subsequently evaluate it on a $G$-transformed test set.
This transformation introduces a distribution shift between $\mathcal{D}_{\text{train}}$ and $G\mathcal{D}_{\text{test}}$.
Therefore, requiring $p(y \mid \mathbf{x}, \mathcal{D}_{\text{train}})$ to perform a zero-shot task on the out-of-distribution test set.
We argue, that this more accurately reflects real-world scenarios.
Under slight abuse of notations, we define $G\mathcal{D}_{\text{test}} \coloneqq \{ \langle \mathbf{x}^\prime, y \rangle \mid \mathbf{x}^\prime = g\mathbf{x}, \langle \mathbf{x}, y \rangle \sim \mathcal{D}_{\text{test}}, g \sim G \}$.
Note that, $G\mathcal{D}_{\text{test}}$ and the vanilla $\mathcal{D}_{\text{test}}$ intersect as the stabilisers of all samples are contained.
For all our experiments, we use $\mathcal{D}_{\text{train}}$ for training and $G\mathcal{D}_{\text{test}}$ during inference.


\paragraph{Max-Confidence Estimation}
We derive an energy-based evaluation from the concept of invariances.
We assume that the function $f_{\mathbf{\theta}}$ is sensitive to $G$, which is true for all neural networks when $G$ is getting more complex.
That is, for any non-trivial element $g$ in $G$ we have $f_{\mathbf{\theta}}(\mathbf{x}) \neq f_{\mathbf{\theta}}(g \mathbf{x})$.
However, based on \cref{def:axioms}, there exists a $g^{-1} \in G$ such that $f_{\mathbf{\theta}}(\mathbf{x}) = f_{\mathbf{\theta}}(g^{-1} g \mathbf{x})$.
Inspired by \citet{Kaba2023} we can determine $g^{-1}$ using a confidence measure $\rho: f_{\mathbf{\theta}}(\cdot) \to \mathbb{R}$, such that
\begin{equation} \label{eq:inverse_approx}
    g^{-1} \leftarrow \text{arg}\max_{h \in G} \rho \left( f_{\mathbf{\theta}}(h g \mathbf{x}) \right).
\end{equation}
This would allow us to undo perturbations induced by real-world measurements. 
An illustration of a confidence surface is depicted in \cref{fig:manifold} (right).
As $\rho$ is the limiting factor of such an approach, we study three different options in the following.

\paragraph{Reducing Prerequisites}
Any such confidence measure relies on a semantically structured latent space \cite{Bengio2012}.
Although a multitude of methods exist to access prediction confidence.
Many of these either pose a prior on the model to learn confidence during training \cite{Moon2020} or require access to the training data to approximate the data manifold \citep{Jake2017, kotelevskii2022}. 
To maintain a minimal level of obstacles, we seek a post-hoc method without any knowledge about the training set.

\paragraph{Energy-induced Confidence}
The probability masses of softmax-classifiers are often spoiled by overconfident posteriors even for out-of-distribution samples \citep{Lee2018, Geifman2018}.
Therefore, the model's energy surface is often used as a straightforward approximation to the confidence surface \citep{Liu2020}.
\citet{Duvenaud2020} showed that Energy-based models \citep{LeCun2006} are implicitly contained in both discriminative and generative models.
This renders Energy-based measures widely applicable.
Following \citet{Duvenaud2020} we determine the total energy of a model $f_{\mathbf{\theta}}$ by 
\begin{equation} \label{eq:energy}
    E_{\mathbf{\theta}}(\mathbf{x}) = -f_{\mathbf{\theta}}(\mathbf{x}) = - \log \sum_y \exp f_{\mathbf{\theta}}(y \mid \mathbf{x}),
\end{equation}
with $p_{\mathbf{\theta}}( \mathbf{x}) \propto \sum_{y^{\prime}} \exp \left(f_\theta(y^{\prime} \mid \mathbf{x})\right)$.
We denote class-wise energies by $E_{\mathbf{\theta}}(y \mid \mathbf{x}) = -f_{\mathbf{\theta}}(y \mid \mathbf{x})$.
Using this relation, we can use any trained classifier and interpret its logit scores as energies.
\citet{Liu2020} showed that the negative energy is a straightforward confidence estimate.

\paragraph{Bayesian-induced Confidence}
Commonly, logits in \cref{eq:boltzmann} are parameterized by a single $\mathbf{\theta}$ obtained by Maximum Likelihood Estimation on $\mathcal{D}_{\text{train}}$.
Instead of such point estimates, Bayesian neural networks \citep{Osband2016} approximate a parameter distribution, hence enabling richer statistical insights, such as confidence estimation.
Unfortunately, exact posterior estimation is known to be intractable due to the integration over the parameter space.
Unlike common approximation schemes \citep{Osawa2019} including ensemble methods \citep{Lakshminarayanan2017}, we only have access to a trained classifier to ease the applicability of \gls*{its}.

To the best of our knowledge, the only method to fulfil this requirement is Monte-Carlo Dropout \citep{Gal2016}. 
Each parameter is endowed with a random process regulating its influence in the forward pass during inference.
Let $\mathbf{\Lambda}_l := \operatorname{diag}(\boldsymbol{\lambda}_l) \sim \text{Bernoulli}(\lambda \in [0, 1])$ be a diagonal matrix of random variables for some layer $l$.
For each inference pass, we sample $\mathbf{\Lambda}_l$ and apply it to the weights as multiplicative noise, $\mathbf{W}_l \leftarrow \mathbf{\Lambda}_l \mathbf{W}_l$.
We can steer the chance of dropout by $1-\lambda$.
This allows us to define an unbiased estimator for the expectation, 
\begin{equation} \label{eq:mcd_expectation}
    \mathbb{E}_{p_{\mathbf{\theta}}(y \mid \mathbf{x})}\left[y\right] 
    = \int y p_{\mathbf{\theta}}(y \mid \mathbf{x}) d y
    \approx \frac{1}{M} \sum_{m=1}^M p_{\mathbf{\theta}}(y \mid \mathbf{x}, \{ \mathbf{\Lambda}_{m,l}\}_{\forall l}).
\end{equation} 
The model's confidence corresponds to the negative total uncertainty, which we can estimate using the entropy,
\begin{align} \label{eq:epistemic}
    \mathbb{H}
    \left[ \mathbb{E}_{p_{\mathbf{\theta}}(y \mid \mathbf{x})} \left[ y \right] \right]
    \approx - \sum_y \mathbb{E}_{p_{\mathbf{\theta}}(y \mid \mathbf{x})} \left[ y \right] \log_2 \mathbb{E}_{p_{\mathbf{\theta}}(y \mid \mathbf{x})} \left[ y \right].
\end{align}

\begin{figure}
    \centering
    \includegraphics[width=0.5\textwidth]{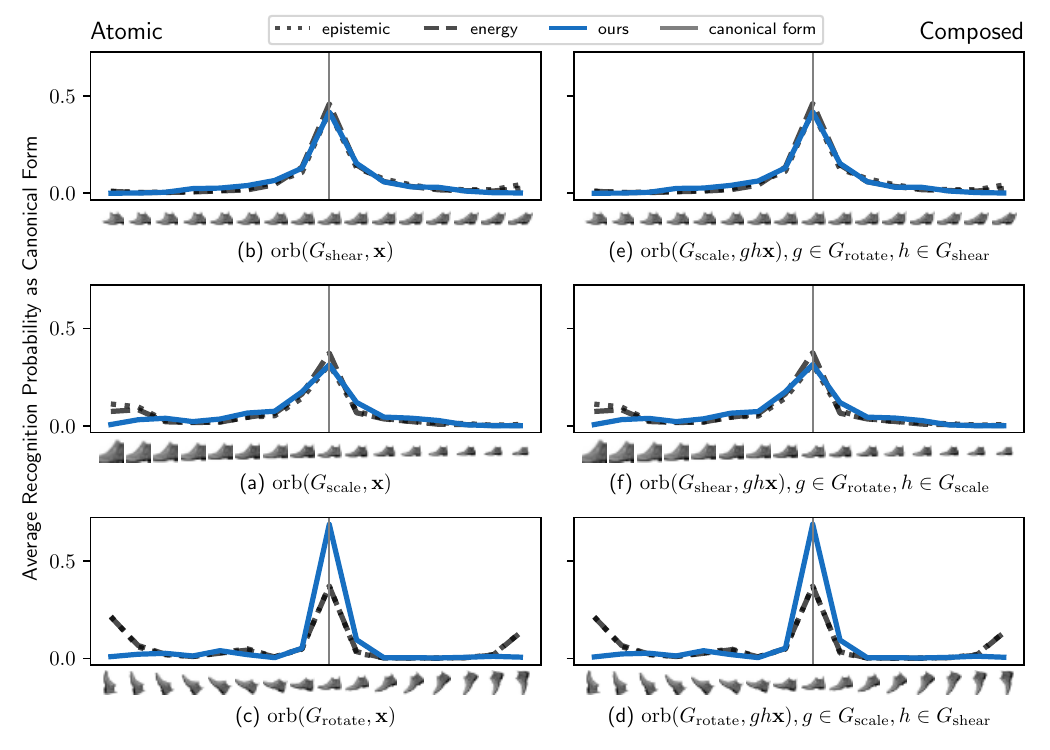}
    \vspace{-1cm}
    \caption{Probability of recognising a transformed sample to be the canonical form. Each evaluation point represents the mean over the FashionMNIST test set.}
    \label{fig:uncertainty}
\end{figure}

\paragraph{Confidence Anomalies}
The confidence measure should concentrate all its mass around the canonical form.
However, we found that both baselines tend to increase their mass towards the domain borders.
This is best observed in the rotation experiment in \cref{fig:uncertainty} (last row).
Here, the spatial alignment bias of FashionMNIST \cite{FashionMNIST} leads to increased confidence for $90^\circ$ rotated images (high activation of horizontal edge detectors). 
Similar confidence bumps appear when scaling up or down a sample until pixelation or disappearance.
Without explicit training or inductive biases, neural networks corrupt the geometry of group orbits during encoding \cite{Arvanitidis2017}.
When traversing the orbit's trajectory, the energy will fluctuate \cite{Li2018}.
As both confidence scores utilise the energy, the confidence will fluctuate likewise.


\paragraph{Group-induced Confidence}
We propose a more stable measure, which does not suffer from these confidence anomalies.
First, we remove noise from the energy by estimating its expected value $\bar{E}_{\mathbf{\theta}}(\cdot)$ using Monte-Carlo Dropout.
This renders the measure stochastic (as the Bayesian-induced confidence), reducing the impact of outliers. 
To further mitigate instabilities, we evaluate local neighbourhoods instead of single point estimates.
This can be achieved by convolving the energy estimate over the orbit with a kernel. 
Lastly, we compute the negative curvature of the resulting energy surface instead of using the energy estimate directly.
This allows us to reduce the confidence mass at the domain borders by using ``nearest" padding, which lowers the curvature at these points.
We approximate the curvature by the second derivative, such that 
\begin{align} \label{eq:group_confidence}
    \rho_G(\mathbf{x}, g) = -\frac{\partial^2}{\partial g^2} \big( \bar{E}_{\mathbf{\theta}} \star K \big)\big(g\mathbf{x}\big),
\end{align}
where $K$ is a Gaussian smoothing kernel to reduce local over-confidence.
We use a generator $\hat{g}$ to generate the finite group $G$.
Therefore, we can evolve $\mathbf{x}$ along $\text{orb}_G(\mathbf{x})$ by incrementing or decrementing $n$ in $\hat{g}^n \mathbf{x}$.
In practice, we can approximate \cref{eq:group_confidence} as follows.
\begin{theorem}[Group Confidence] \label{thm:group_confidence}
    Let $\phi_{\bar{E}_{\mathbf{\theta}}, K}(\mathbf{x}, g) := \big( \bar{E}_{\mathbf{\theta}} \star K \big)\big(g\mathbf{x}\big)$.
    The group-induced confidence defined in \cref{eq:group_confidence} can be approximated by
    \begin{align}
    &\rho_G(\mathbf{x}, g) = -\lim_{|G| \to \infty} \frac{1}{\| \hat{g} - \hat{g}^2 \|^2} \\ 
    &\bigg( \phi_{\bar{E}_{\mathbf{\theta}}, K}(\mathbf{x}, \hat{g} \circ g) + \phi_{\bar{E}_{\mathbf{\theta}}, K}(\mathbf{x}, \hat{g}^{-1} \circ g) - 2\phi_{\bar{E}_{\mathbf{\theta}}, K}(\mathbf{x}, g) \bigg)\nonumber.
    \end{align}
\end{theorem}
Find the proofs in \cref{proof:orbit} and \cref{proof:group_confidence}.

A comparison of all three measures is depicted in \cref{fig:uncertainty}.
Our group-induced confidence either surpasses both baselines or achieves on par results. 
\section{Inverse Transformation Search}
\label{sec:methodology}

\begin{figure}[t]
\begin{center}
\centerline{\includegraphics[width=\columnwidth]{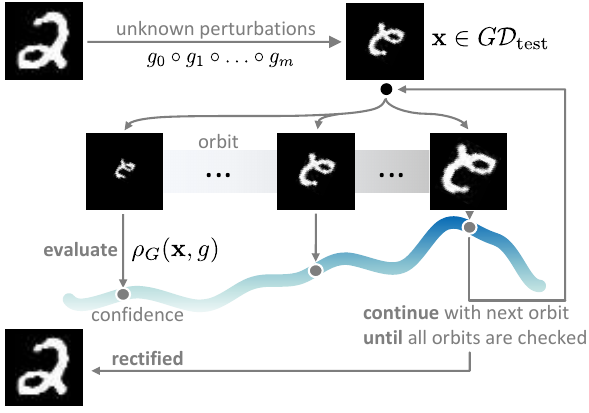}}
\caption{Our proposed \glsfirst*{its} on a perturbed image. In each level of the search tree, the orbit of one subgroup is evaluated. All samples are evaluated using \cref{eq:group_confidence} and a pre-trained classifier.
}
\label{fig:algorithm}
\vspace{-1cm}
\end{center}
\end{figure}

\begin{figure*}[t]
\begin{center}
\centerline{\includegraphics[width=\textwidth]{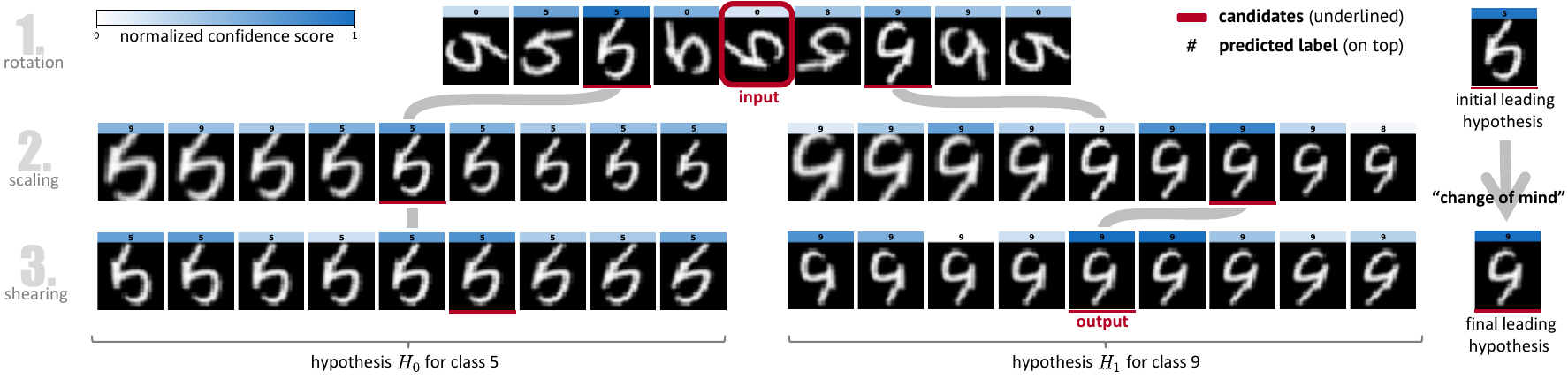}}
\vspace{-0.5cm}
\caption{An example iteration of our algorithm with two hypotheses $H_0$ and $H_1$, and three generated subgroups, each with a cardinality of $|G|=9$. The first layer rotates the input, the second layer scales it, and the last layer shears it.
}
\label{fig:algorithm_example}
\vspace{-0.5cm}
\end{center}
\end{figure*}

An overview and an example of our proposed search procedure are illustrated in \cref{fig:algorithm}.
In the following, we spell out the most important aspects of it with more details in \cref{sec:implementation_details} and \cref{sec:theory} (including pseudocode).

\paragraph{Transforming Images}
An image $\mathbf{x}$ is a collection of pixel vectors containing spatial and colour information.
Spatial transformations are modifications of the location a pixel vector is pointing to.
To resample the transformed image, colour values are interpolated between the modified pixel positions.
This procedure can be efficiently executed in parallel on the GPU \cite{jaderberg2016spatial}. 

\paragraph{Forming Search Levels}
We deal with problem scenarios, where samples are perturbed by a some unknown affine transformation $g$ during test-time.
We assume that $g$ decomposes into various subgroup transformations, $g = g_0 \circ g_1 \circ \ldots \circ g_m$.
We propose an iterative search procedure, which aims to eliminate one $g_k$ at every step using its inverse $g_k^{-1}$.
This heuristic (greedy) approach reduces the complexity of the problem significantly.
However, this comes with the assumption that either the order of transformation is known, which is unlikely in real-world settings, or that we deal with Abelian groups.
We show that the restrictions imposed by the latter can be attenuated, as only a sub-sequence needs to be Abelian.
    
\begin{theorem}[Group Elimination] \label{thm:group_elimination}
Let $s = g_0 \circ g_1 \circ \ldots \circ g_m$ be a finite sequence of group elements with $g_0 \in G_0, g_1 \in G_1, \ldots, g_m \in G_m$ and $m \in \mathbb{N}$.
Further, let $g_k^{-1}$ be the inverse of $g_k$ in the sequence for some $k \in [0, m]$ and $s_{\setminus g_k}$ being $s$ without $g_k$.
We have, $g_k^{-1}s = s_{\setminus g_k}$ if all elements in the sub-sequence $s_{0:k-1}$ and $g_k^{-1}$ are elements of an Abelian group.
\end{theorem}

The proof can be found in \cref{proof:group_elimination}.

To cope with Lie groups, we generate subgroups of size $|G|$ using generators as already used in our group-induced confidence measure (see \cref{thm:group_confidence}).
In that light, each layer comprises $|G|$ elements (as a finite set of equidistant samples along the orbit).
This allows us to express transformations by an exponential $\hat{g}^n$ with $n \in \mathbb{N}$.
This renders the evaluation in each layer tractable, as only a finite set of points needs to be evaluated.\footnote{All non-cyclic groups, like translations, would at some point shift the image signal outside the visible pixel grid. In practice, we consider only a reasonably bounded set of transformations.}

\begin{figure}
    \centering
    \includegraphics[width=\columnwidth]{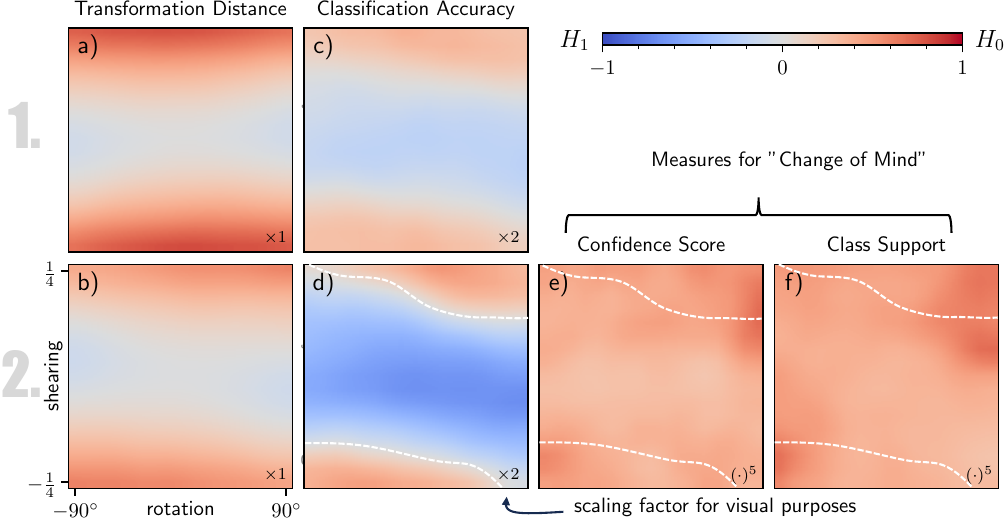}
    \vspace{-0.5cm}
    \caption{Evolution of $H_0$ and $H_1$ in a two-layer search tree (rotation in the top row and shearing in the bottom one) on Fashion-MNIST.
    Red areas indicate an average superiority of $H_0$, while blue areas indicate a superiority of $H_1$.}
    \label{fig:hypotheses}
    \vspace{-0.5cm}
\end{figure}

\paragraph{Hypothesis Testing}
We define a hypothesis $H_i$ as a trajectory down the search tree.
Humans use selective hypothesis testing \cite{Sanbonmatsu1998} when the perceived visual stimulus has high uncertainty.
This includes evaluating multiple hypotheses until a final decision is made.
We simulate this behaviour by traversing the search tree $k$ times in parallel.
This is done by updating a collection of candidate hypotheses $\{ H_0, H_1, \ldots, H_{k-1}\}$.
After the first expansion and evaluation step, we select the $k$ best candidates ordered by confidence.
Each further expansion and evaluation step is done for all $H$.
The $k$ best are selected from the cumulative candidate set, bounding each level of the tree to $k$ children. 
Find an example of such a $k$-ary tree in \cref{fig:algorithm_example}.

Rather than assuming independent trajectories, we pose a linking condition between them to avoid redundant hypotheses.
For instance, a rotated image of a $5$ might share many features with a $9$.
If the confidence is high, a significant amount of runs would refine this hypothesis until the end, resulting in a one-sided perspective.
We improve the hypothesis testing by allowing only one hypothesis per label to be further refined.

\paragraph{Change of Mind}
In theory, such a dynamic search process over multiple hypotheses would allow the model to "change its mind".
That is, rejecting $H_0$ (the initially dominating hypothesis) and instead accepting some other $H$ for the final incumbent (output of the inference process).
\cref{fig:algorithm_example} already provides an example for such phenomena, which we investigate more thoroughly in \cref{fig:hypotheses}.
We constructed a regular grid of rotated and sheared Fashion-MNIST test sets.
Each point in \cref{fig:hypotheses} is the average performance of a \gls*{cnn} with \gls*{its} (see \cref{sec:change_of_mind} for more details).
Colour changes from layers 1 to 2 indicate that the initial hypothesis is on average superseded by the other.
To maximise the performance outcome (we focus on classification accuracy), we need to substitute the leading hypothesis $H_0$ with its alternative $H_1$ in some regions.
Without a ground-truth signal (during test time), we need a measure to determine when to swap hypotheses.
In our experiments, we used class support, which is the number of occurrences of the predicted class over the entire tree.
However, this indicator has only minor advantages over the confidence score (see \cref{fig:hypotheses} e,f).
We leave this promising avenue for future research.

\paragraph{Complexity and Runtime}
We have a fixed-width $k$-ary tree, where the number of elements in the first level is $n := |G|$ and in all consecutive levels $kn$.
As all evaluations are performed in parallel, $n$ does not affect the runtime. 
For our implementation, the Monte Carlo estimates are performed sequentially, scaling the complexity by $M$.
This gives $\mathcal{O}(Mk^d)$, where $d$ denotes the depth of the tree.
We quantified the runtime cost in \cref{sec:runtime}.

\paragraph{Application Domains}
Given the runtime costs, we argue that our inference strategy is particularly interesting for applications where it is worth trading runtime against robustness.
This includes medical applications, like X-ray classification \cite{Ghamizi2023}; analysis of satellite or drone imagery, like land suitability prediction for agriculture \cite{Mehrjardi2020}, monitoring crop health \cite{Inoue2020}, or identifying deforestation \cite{Andrade2020}; classifying celestial objects in telescopic images \cite{Gomez2020} or analysing protein structures in biology from microscopic images \cite{Moen2019}.

\section{Experiments}
\label{sec:experiments}

In each classifier, there is potential for pseudo-invariance, which we just have to activate.  
In that light, all investigations are zero-shot performance tests with out-of-distribution data (see \cref{sec:confidence}).
We split the vanilla datasets into disjunct training, validation, and test sets.
We always employ the vanilla training set to fit the model and validate it on the vanilla validation set.
All performance measures are subsequently evaluated on the perturbed test set $G\mathcal{D}_{\text{test}}$, where $G$ is specified in each experiment.

If not further specified, we used zero-padding to define areas outside the pixel space $\Omega$, bilinear interpolation, and a group cardinality of $n=17$.
For our experiments on (Fahsion)-MNIST and GTSRB we used two \gls*{cnn}-backbones simple as the LeNet architecture \cite{LeNet} (see \cref{sec:model_architectures}).
These models are trained with the AdamW optimizer \cite{AdamW} using default parameters.
We minimised the negative log-likelihood using ground-truth image labels.
We used a learning rate of $5e-3$, $3$ epochs for MNIST, $5$ epochs for Fashion-MNIST, $10$ epochs for GTSRB and mini-batches of size $128$.
In all MNIST experiments, we removed the class $9$ samples, as this introduces a class-superposition for $180^\circ$ rotations.
More details can be found in our publicly available source code.\footnote{\texttt{www.github.com/johSchm/ITS}}



\subsection{Zero-Shot rotated Image Classification}
\label{sec:zero-shot}

\begin{figure}
    \centering
    \includegraphics[width=\columnwidth]{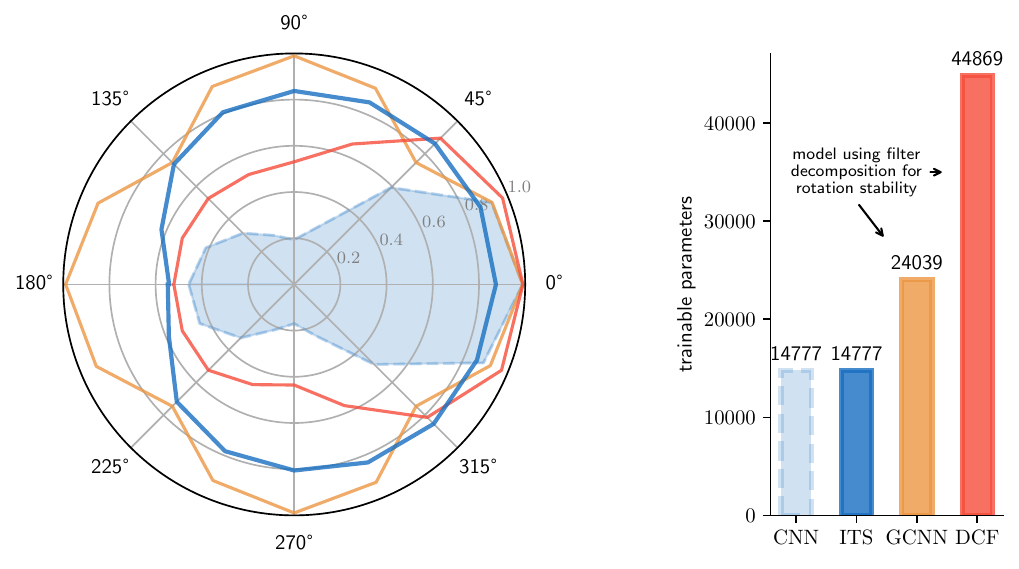}
    \caption{Test accuracy on rotated MNIST with fixed angles.}
    \label{fig:rotation_mnist}
\end{figure}

We trained a \gls*{cnn}, a \gls*{gcnn} \cite{cohen2016group} and a RotDCF \cite{RotDCF} on the vanilla (canonical) MNIST.
Both baselines use Filter Decomposition to gain $p4$-equivariance and $p8$-equivariance, respectively.
We evaluated their zero-shot performances on the rotated test sets with results shown in \cref{fig:rotation_mnist}.
Such a model would be robust to centre-rotations.
Just by exploiting the confidence surface of the vanilla (shallow) baseline \gls*{cnn} our method boosts its robustness significantly.
One can also see, that the rotation-equivariant \gls*{gcnn} outperforms our methods in the majority of test angles.
However, such models require hard-coded inductive biases tailored to some specific groups\footnote{Developing inductive biases for all possible transformations is impossible due to the vastness of possible transformations.}.
Our method does not pose any such requirements in the hypothesis space, also allowing for smaller model complexity.
We enable pseudo-equivariance to more transformations, which is shown in the subsequent experiments.



\subsection{Zero-Shot Affine Canonicalisation}
\label{sec:exp_affine_canonical}

\begin{table*}[t]
\caption{Classification accuracies ($\uparrow$) and Frobenius distances ($\downarrow$) between the ground truth and the predicted transformation matrix.
Mean and standard deviations are estimated over $25$ test set runs with random rotation, scaling, and shearing parameters.}
\label{tab:performance}
\vskip 0.15in
\begin{center}
\begin{small}
\begin{sc}
\begin{tabular}{lccccccc}
\toprule
Dataset & CNN & STN & STN$^3$ & DTN & ETN & ITS$^3$ & ITS$^5$ \\
\midrule
MNIST    & 50.15$\pm$9.29 & 49.78$\pm$9.13 & 13.85$\pm$4.05 & 48.61$\pm$8.99 & 56.59$\pm$9.87 & 88.89$\pm$13.86 & \textbf{89.81$\pm$13.95} \\
F-MNIST  & 25.91$\pm$5.93 & 24.51$\pm$5.81 & 12.42$\pm$3.73 & 24.11$\pm$5.71 & 26.10$\pm$5.98 & \textbf{38.47$\pm$7.77} & 37.58$\pm$7.56 \\
GTSRB    & 32.01$\pm$6.59 & 31.46$\pm$6.55 & 2.78$\pm$1.69 & 28.21$\pm$6.23 & 40.02$\pm$7.99 & 66.44$\pm$11.48 & \textbf{67.09$\pm$11.53} \\
\midrule
MNIST    & 23.03$\pm$2.48 & 25.43$\pm$2.72 & 23.61$\pm$2.54 & 31.02$\pm$3.31 & 26.10$\pm$2.83 & \textbf{18.80$\pm$2.22} & 18.98$\pm$2.32 \\
F-MNIST  & 23.03$\pm$2.80 & 25.96$\pm$3.14 & 23.61$\pm$2.87 & 35.16$\pm$4.22 & 34.02$\pm$4.19 & \textbf{20.25$\pm$2.59} & \textbf{20.30$\pm$2.56} \\
GTSRB    & 23.04$\pm$2.85 & 18.15$\pm$2.23 & 23.62$\pm$2.92 & 34.61$\pm$4.23 & 21.08$\pm$2.61 & \textbf{17.87$\pm$2.26} & \textbf{17.93$\pm$2.27} \\
\bottomrule
\end{tabular}
\end{sc}
\end{small}
\end{center}
\vskip -0.1in
\end{table*}

\begin{figure}
    \centering
    \includegraphics[width=\columnwidth]{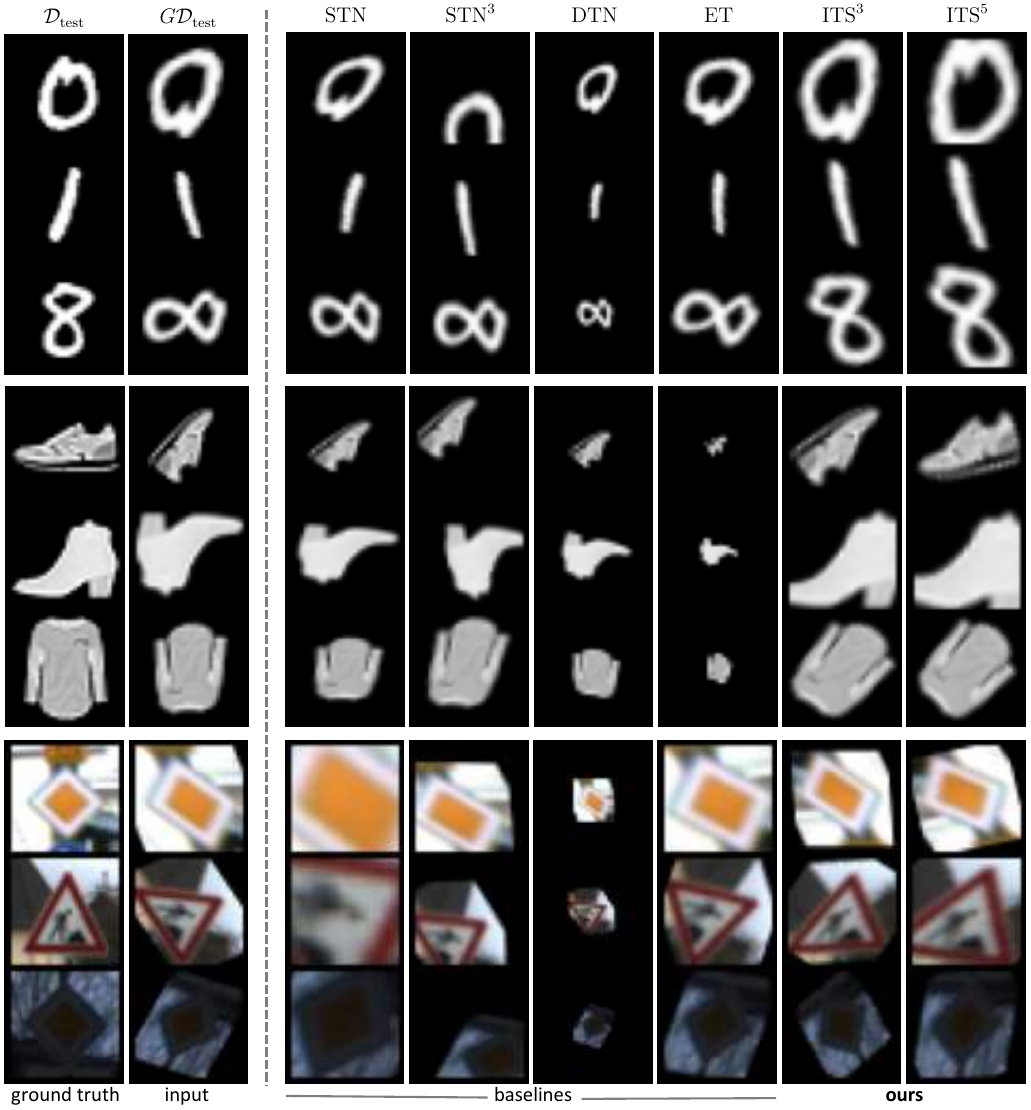}
    \vspace{-0.5cm}
    \caption{Three test set samples of MNIST (top), Fashion-MNIST (middle), GTSRB \cite{GTSRB} (bottom), which are transformed by a composed spatial transformation of rotation $[-\pi, \pi]$, scaling $[-\frac{1}{4}, \frac{1}{4}]$ and shearing $[-\frac{1}{4}, \frac{1}{4}]$. All models use the same \gls*{cnn} backbone, are trained on $\mathcal{D}_{\text{train}}$ and are tested on $G\mathcal{D}_{\text{test}}$. \gls*{its} is used with $3$ and $5$ levels and compared to a \gls*{stn} and a \gls*{dtn} with a single and $3$ transformation layers, respectively.}
    \label{fig:reconstructions}
    \vspace{-0.5cm}
\end{figure}

Similar to \glspl*{stn}, we can undo affine transformations (i.e., canonicalize inputs).
We compare \gls*{its} to a \gls*{stn} \cite{jaderberg2016spatial} with one and three transformation layers using inverse compositional layers \cite{Lin2017}, a \glsfirst*{dtn} \cite{Detlefsen2018}, and a \glsfirst*{etn} \cite{tai2019equivariant}.
To study the impact of depth, we contrasted a $3$- and a $5$-layer \gls*{its}.
We used a layer stack of rotation + scaling + shearing + (rotation + scaling).
We refer to these as \gls*{its}$^3$ and \gls*{its}$^5$, respectively.
\cref{tab:performance} shows the quantitative results of our experiments.
The associated qualitative results as reconstructions (after applying the predicted inverse transformation) can be found in \cref{fig:reconstructions}.

We found all baselines incapable of performing affine canonicalisations outside the training data distribution.
This makes sense, as their localisation networks (regressing the transformation parameters) learn pixel-to-transformation mappings from training data. 
They are unable to generalise - particularly to rotations, which are mostly present during testing.
The \gls*{stn} and the \gls*{dtn} estimate the composed transformation matrix in one forward pass. 
They need to extrapolate from nearly no rotations during training to $360^\circ$ rotations during testing.
As the rotation manifold is isomorphic to the unit circle ($\cos(\theta), \sin(\theta)$), this non-linearity needs to be learned implicitly.
The \gls*{etn} on the other hand uses canonical coordinates, which linearise the relationship between rotations.

While \gls*{its} does not always lead to correct results, it works best among the contestants.
In \cref{tab:performance} we also see that for most cases the Frobenius distance is lowest, too.
Interestingly, the distance to the ground truth matrix is lowest for \gls*{its}$^5$ while \gls*{its}$^3$ achieves higher accuracies. 
High Frobenius distance and high accuracy would indicate that the classifier got an incorrectly rectified sample but is still able to predict the correct class.

\subsection{Scaling up to ImageNet} \label{sec:siscore}

\begin{table*}
  \centering
  \caption{Average acc@1 and acc@5 over the SI-Rotation test set using a \gls*{vit}-B16 pre-trained on ImageNet. The vanilla performance is compared to a rotation and scale augmented fine-tuning, training on foreground objects with and without annotated segmentation masks (AS) and our \gls*{its}. We used two acc@5 estimates for \gls*{its}: over $H_0$'s categorical distribution and over the top-1 predictions of all $H$.}
  \vspace{0.1cm}
  \begin{tabular}{p{\columnwidth}lll}
    \toprule
    \centering SI-Rotation Test Set Samples & ViT & acc@1 & acc@5 \\
    \midrule
    \multirow{5}{*}{\begin{minipage}{0.5\columnwidth}\includegraphics[width=2.\columnwidth, height=.5\columnwidth]{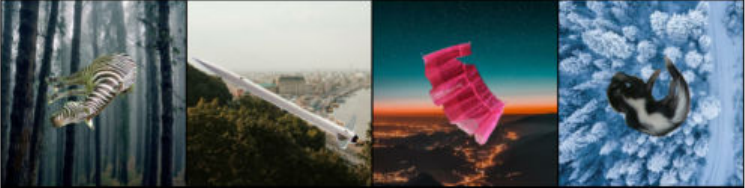}\end{minipage}} & vanilla & 38.5 & 58.5 \\
    & Rotation + Scale Augm. & 41.0 \textcolor{Gain}{(+2.5)} & 63.2 \textcolor{Gain}{(+4.7)} \\
    & \cite{Chefer2022} & 44.8 \textcolor{Gain}{(+6.3)} & 65.2 \textcolor{Gain}{(+6.7)} \\
    & \cite{Chefer2022}+AS & 46.2 \textcolor{Gain}{(+7.7)} & 67.0 \textcolor{Gain}{(+8.5)} \\
    & \gls*{its} \emph{(ours)} & \textbf{49.0 \textcolor{Gain}{(+10.5)}} & 68.7 \textcolor{Gain}{(+10.2)} \\
    & & & \textbf{70.9 \textcolor{Gain}{(+12.4)}} \\
    \midrule
    \bottomrule
  \end{tabular}
  \label{tab:siscore}
\end{table*}

\begin{figure*}
    \centering
    \includegraphics[width=\textwidth]{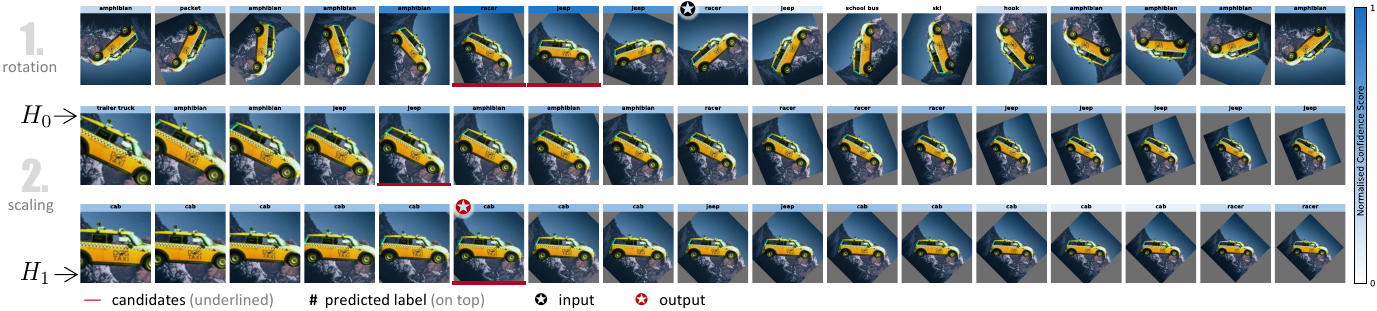}
    \caption{SI-Rotation sample, showing a tilted image of a cropped cab (ground-truth label) pasted on a mountain-view background.
    We applied \gls*{its} with rotation $[-\pi, \pi]$ (first row) and isotropic-scaling $[-\frac{1}{2}, \frac{1}{2}]$ (second and third row) with two hypotheses ($H_0$ and $H_1$).}
    \label{fig:rotation_example}
\end{figure*}

Si-Score (short SI) \cite{Djolonga2021} is a synthetic vision dataset for robustness testing, comprising semantically masked ImageNet \cite{Imagenet2015} objects pasted on various backgrounds.
The foreground objects are either arbitrarily (horizontally and vertically) translated (SI-Location), arbitrarily rotated (SI-Rotation), or arbitrarily scaled (SI-Size).
We evaluate \gls*{its} using a \glsfirst*{vit}-B16 \cite{Dosovitskiy2020}.
The Vision Transformer is pre-trained and validated on ImageNet and tested on Si-Score. 
We trained the model with and without data augmentation using random rotations $[-\pi, \pi]$ and scales $[-\frac{1}{4}, \frac{1}{4}]$.

We found significant improvements on SI-Rotation as denoted in \cref{tab:siscore}.
To some extent this was expected as our confidence measure is highly sensitive in detecting canonical rotations (see \cref{fig:uncertainty}).
Surprisingly, however, our method even outperformed augmented fine-tuning and explicit training on the foreground object \cite{Chefer2022}.
An example is depicted in \cref{fig:rotation_example}.
This shows the brittleness of ImageNet-trained models to spatial transformed inputs \cite{Engstrom2019} (especially to rotations), as the label is changing frequently along the orbits.
An exception is $H_1$ (third row), where the correct label is preserved over almost the entire orbit of isotropic-scalings. 
This property is studied in SI-Size.

On SI-Size \gls*{its} maintains the performance of the baseline.
We hypothesise that location and size are properties that have high variance in the vanilla ImageNet already.
This puts the test performance on a high level without requiring any additional mechanisms.
Equipping the backbone with the ability to translate the input query with two degrees of freedom causes it to focus on the wrong object.
This behaviour is similar to attention but with a fixed centre bias of the backbone and a spatial translation of the signal instead.
However, \gls*{its} does not get any feedback (gradient) if the object of focus is correct.
This leads to slightly degenerated performance on SI-Location (see \cref{sec:siscore_cont}). 

Furthermore, we tested two variants to obtain acc@5: 
(i) We computed $H_0$'s categorical distribution and picked the top-5 predictions.
(ii) We used the top-1 predictions of each of the five hypotheses ($H_0 \ldots H_4$).
A higher score for the former would indicate that on average $H_0$ followed the correct class hypothesis.
Whereas, when the latter is higher, one of the alternative hypotheses ($H_1 \ldots H_4$) contained the true class more frequently than $H_0$.
Unfortunately, our experiments do not show a consistent pattern of which strategy yields higher accuracies.
We leave further investigations to future work.

\section{Conclusion}
\label{sec:conclusion}

\glsfirst*{its} equips any vanilla image classifier with pseudo-invariance to various spatial transformations during inference.
This increases its robustness to spatially perturbed inputs without requiring extra training data.
We activate this potential, which lies hidden in any classifier, without requiring any inductive biases or augmented training.
This is achieved by an incremental search for the inverse perturbation.
At the canonical form, the features of the transformed input match the features known from the training data.
At these points, the model is most confident to classify this sample properly.
We successfully showed the superiority of our methods for zero-shot image classification tasks.

\subsection{Limitations}

Mitigating the following three limitations should be the focus of future works: 
\emph{(1) False Hypotheses:} Heavily perturbed images can have multiple proper hypotheses.
See the example in \cref{fig:hypotheses}, where both labels ($5$ or $9$) seem reasonable.
As \gls*{its} uses the representation space of the trained classifier, its performance depends on the quality of these representations.
The model might detect features in transformed signals, which lead to false hypotheses. 
Even worse, these hypotheses might be reinforced down the search tree.
See \cref{fig:focus-example1} for an example of such behaviour.
\emph{(2) Abelian Assumption:} In \cref{thm:group_elimination}, we showed that only a subsequence of group actions need to be Abelian. 
Although, this alleviates the impact of the assumption, the issue is not solved.
\emph{(3) Runtime Cost:} Due to the complexity of the search process (see \cref{sec:methodology}) the walltime increases significantly during inference (see \cref{sec:runtime}). 
In many setups, like video understanding \cite{Schmidt2021}, this renders real-time applications intractable.

\subsection{Future Work}
Given the energy field of the target model, any post-hoc optimisation method can be applied to maximise the confidence of a test sample over a transformation group.
This search can be done over the entire group space, not only over subgroups (as used here) tackling the issue with Abelian assumption.
This opens up a wide range of algorithms, including for example simulated annealing, which is used in a wide range of domains \cite{Landau2005, Schmidt2024}.
As such algorithms are applicable to Quantum-Computers, they hold the potential to eliminate the runtime issue.

\section{Impact Statement}
This paper presents work whose goal is to advance the field of Machine Learning. There are many potential societal consequences of our work, none of which we feel must be specifically highlighted here.

\section{Acknowledgement}
Special thanks to all our lab members for the extremely valuable feedback and support during the process.
Additionally, we thank the anonymous reviewers for their insightful comments that helped improve this work.

{
    \small
    \bibliographystyle{icml2024}
    \bibliography{main}
}

\clearpage
\setcounter{page}{1}
\appendix

\section{Implementation Details}
\label{sec:implementation_details}

If not otherwise stated in the experiment descriptions, we used a group cardinality of $n=17$.
The choice of odd numbers is due to our algorithmic implementation.
We divide the parameter space into $n$ equally distant chunks.
An odd $n$ ensures that the identity is included.

\subsection{Search Algorithm}
Supplementary to the descriptions and illustrations in \cref{fig:algorithm} and \cref{fig:algorithm_example}, we provide the pseudocode of our proposed algorithm in \cref{alg:method}.
The complexity of the search increases if more than one hypothesis is maintained.
We refer the reader to our publicly available source code for more details.\footnote{\texttt{www.github.com/johSchm/ITS}}


\begin{algorithm}[tb]
   \caption{Pseudocode for \gls*{its} with one hypothesis $H$.}
   \label{alg:method}
\begin{algorithmic}
   \STATE {\bfseries Input:} $\mathbf{x} \sim G\mathcal{D}_{\text{test}}$
   \STATE $\rho_{\text{primal}} = 0$ and $h = \{e\}$
   \FOR{$G$ {\bfseries in} $\{G\}$}
   \STATE $\rho, g = \arg \max_{g \in G} [\rho_G(\mathbf{x},g \circ h)]$
   \IF{$\rho > \rho_{\text{primal}}$}
   \STATE $\rho_{\text{primal}} \leftarrow \rho$
   \STATE $H \leftarrow p_\theta(y \mid g \circ h \mathbf{x})$
   \STATE $h \leftarrow g \circ h$
   \ENDIF
   \ENDFOR
   \STATE {\bfseries Output:} $h$
\end{algorithmic}
\end{algorithm}

\subsection{Model Architectures}
\label{sec:model_architectures}

\paragraph{Small-Sized CNN}
This is used for experiments on MNIST and Fashion-MNIST.
This model comprises three 2D-convolutional layers with $5 \times 5$ kernels and $8$, $16$, and $16$ channels.
We add GeLU non-linearities \cite{GELU} after each convolution layer.
2D Max-Pooling with $2 \times 2$ kernels is added after the second and the third convolutional layer.
The final stack of feature maps is flattened into a vector and further compressed to the number of classes by two linear layers.
After the first linear layer, we also add a GeLU non-linearity and a dropout-layer with $50\%$ of dropout.
From here we read out the logit scores.

\paragraph{Large-Sized CNN}
This is used for experiments on GTSRB.
This model comprises three 2D-convolutional layers with kernel sizes of $5$, $3$ and $1$ and $100$, $150$, and $250$ channels.
We add GeLU non-linearities \cite{GELU}, 2D Max-Pooling with $2 \times 2$ kernels, Batch Normalisations \cite{BatchNorm} and Dropout after each convolution layer.
The final stack of feature maps is flattened into a vector and further compressed to the number of classes by two linear layers.
After the first linear layer, we also add a GeLU non-linearity and a dropout-layer with $50\%$ of dropout.
From here we read out the logit scores.

\section{Further Related Works}

\subsection{Test-time Domain Adaptation}
In our work, we model a canonicalisation function by a search process. Canonicalisation \cite{Kaba2023, Mondal2023} is build on first principles and supported by the mathematical guarantees of group theory. This is highly distinctive to learnable test-time adaptions (TTAs) \cite{Wang2021, Wang2022, Wu2024}. We do not need to modify the parameters of the neural network, we can even treat the model as a black box only having access to the logit-scores. Source-free TTAs often suffer from error accumulations and catastrophic forgetting \cite{Wang2022}. For our method, the latter is avoided a priori, as at no point the neural network and its weights are modified. Similar to \citep{Wang2022}, we use weight averages (approx. Bayesian) to mitigate error accumulations.

\section{Further Experiments}

\paragraph{Remark regarding Image-Backgrounds}
Spatial transformations can introduce background artefacts for datasets without mono-colour backgrounds, like ImageNet. There exist only solutions for a few quotient groups (like rotations \cite{Tuggener2023}) not for the entire affine group (as considered in this work). Fortunately, this is only an issue during training as the model might exploit these artefacts, which does not apply for any conducted experiment. Either all models are trained on the vanilla dataset (as in \cref{sec:zero-shot} and \cref{sec:exp_affine_canonical}) or are trained with artefacts (baselines in \cref{sec:siscore}) but tested on an artefact-free dataset (here SiScore).

\subsection{Runtime}
\label{sec:runtime}

\begin{figure}
    \centering
    \includegraphics[width=0.8\columnwidth]{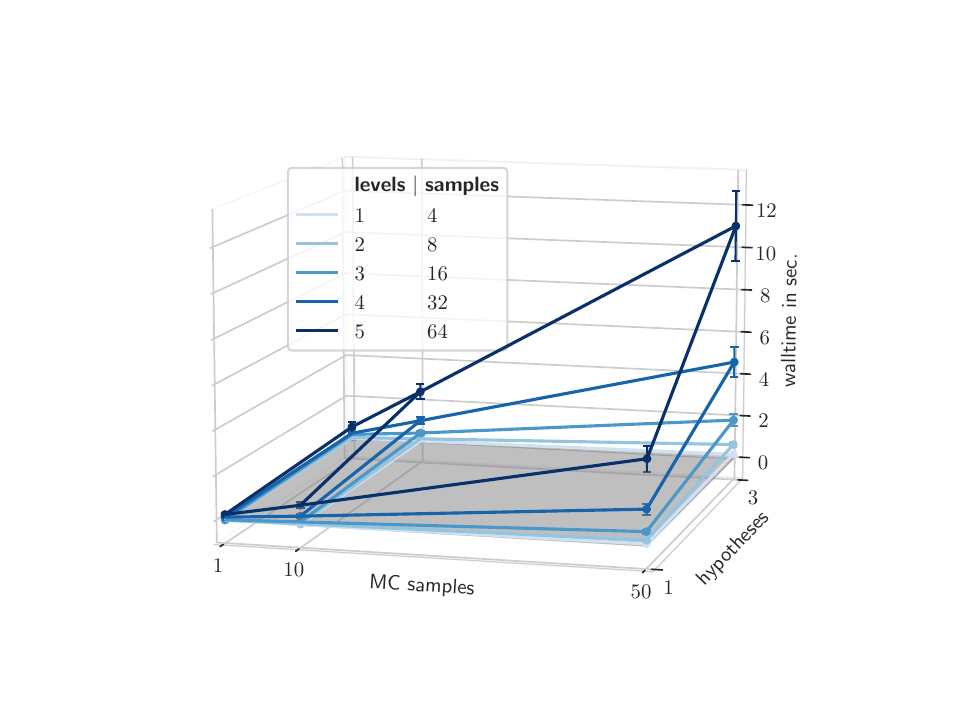}
    \caption{Walltime of \gls*{its} (shades of blue) with a \gls*{cnn} backbone (gray) on a transformed MNIST test set (standard deviation shown as error bars).}
    \label{fig:runtime}
\end{figure}

All experiments are performed on an Nvidia A40 GPU (48GB) node with 1 TB RAM, 2x 24core AMD EPYC 74F3 CPU @ 3.20GHz, and a local SSD (NVMe).
The software specifications of our implementations can be found in our open-sourced code.
\cref{fig:runtime} shows how the runtime behaves under different hyperparameter settings compared to the pure backbone.
The most significant factor is the depth of the search tree.
In \cref{sec:exp_affine_canonical} we found that the performance does not notably increase with more hypotheses.
The number of Monte-Carlo (MC) dropout steps, stabilises the predictions by reducing temporal confidence anomalies.
Although, we used $50$ MC samples for our experiments, the impact on the performance is minor.
Another factor is the number of hypotheses tested.
These hypotheses are processed in parallel but due to the unique class condition (see \cref{sec:methodology}), the runtime scales with the number of hypotheses.
Empirically, we found that three to five produces the best results.

\begin{figure}
    \centering
    \includegraphics[width=\columnwidth]{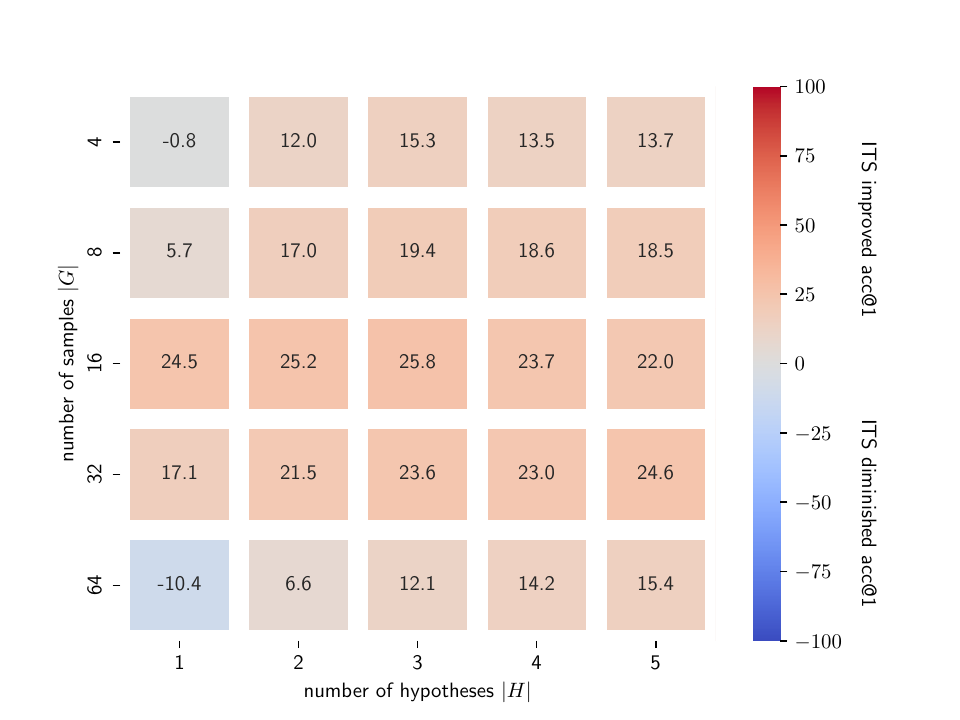}
    \vspace{-0.5cm}
    \caption{Average acc@1 improvement on the randomly rotated and scaled MNIST test set compared to the baseline (without ITS) performance.}
    \label{fig:samples_vs_hypotheses}
    \vspace{-0.5cm}
\end{figure}

\subsection{Hyperparameter Tuning for Accuracy}
We continue the previous investigation by studying the influence of these hyperparameters on the accuracy.
Increasing the number of MC samples stabilises predictions but also reduces exploration. When increasing the MC samples from $1$ to $10$ on affine-transformed Fashion-MNIST the acc@1 increases $9.46\pm1.14$ points (mean and standard deviation over $10$ runs). Increasing the MC samples from $10$ to $50$ leads to an acc@1 increase of only $1.38\pm0.47$.

In \cref{fig:samples_vs_hypotheses}, we illustrated the performance improvements with respect to the number of samples per layer and the number of hypotheses.
Interestingly, the performance does not increase (or stagnate) when increasing either value.
It seems that the middle ground around $|H|=3$ and $|G|=16$ results in maximum improvements compared to the vanilla model (without ITS) prediction.
Due to the unique-label constraint of each hypothesis (see \cref{sec:methodology}) many hypotheses with initial low confidence are pursued.
This might lead to catastrophic reinforcement of many false hypotheses, obtaining higher confidences down the line.

\subsection{Change of Mind}
\label{sec:change_of_mind}

To compute the results in \cref{fig:hypotheses}, we used the following measures:
We estimated the transformation distance between the predicted matrix, $\mathbf{T}_{H_0}$ or $\mathbf{T}_{H_1}$, and the ground-truth matrix, $\mathbf{T}_{\mathrm{GT}}$.
Let $d(\mathbf{T}_{H_i}, \mathbf{T}_{\mathrm{GT}}) := \| \mathbf{T}_{H_i} - \mathbf{T}_{\mathrm{GT}} \|_F$ for $i \in \{ 0,1 \}$.
To ease comparability, we used a cross-normalisation to relate the outcomes.
Let $\mathcal{T}_{G\mathcal{D}_{\text{test}}}[H_i]$ be the set of predicted transformation matrices of $H_i$ for all samples in $G\mathcal{D}_{\text{test}}$.
Then, the minimum and maximum distance can be obtained by
\begin{align}
    d_{\text{min}} &:= \min_{\mathbf{T} \in \mathcal{T}_{G\mathcal{D}_{\text{test}}}[H_0] \cup \mathcal{T}_{G\mathcal{D}_{\text{test}}}[H_1]} d(\mathbf{T}, \mathbf{T}_{\mathrm{GT}}), \\
    d_{\text{max}} &:= \max_{\mathbf{T} \in \mathcal{T}_{G\mathcal{D}_{\text{test}}}[H_0] \cup \mathcal{T}_{G\mathcal{D}_{\text{test}}}[H_1]} d(\mathbf{T}, \mathbf{T}_{\mathrm{GT}}).
\end{align}
We can min-max cross-normalise each distance by
\begin{equation}
    \bar{d}(\mathbf{T}_{H_i}, \mathbf{T}_{\mathrm{GT}}) = \frac{d(\mathbf{T}_{H_i}, \mathbf{T}_{\mathrm{GT}}) - d_{\text{min}}}{d_{\text{max}} - d_{\text{min}}} \in [0,1].
\end{equation}
We are interested in the difference between both predictions,
\begin{equation}
    \bar{d}(\mathbf{T}_{H_0}, \mathbf{T}_{\mathrm{GT}}) - \bar{d}(\mathbf{T}_{H_1}, \mathbf{T}_{\mathrm{GT}}).
\end{equation}
This is the measure used to produce each point in \cref{fig:hypotheses} \emph{(a,c)}.
The construction of the classification accuracy plots in \cref{fig:hypotheses} \emph{(b,d)} is done similarly.
We counted the correct top-1 class-predictions of $H_0$ and $H_1$ over $G\mathcal{D}_{\text{test}}$.
Again, cross-normalised the results and computed the difference.
\\
The confidence score differences and the class support differences are only plotted for the last layer.
This is done as the re-ranking of the hypotheses is only required at the end of the search procedure for the final read-out of the incumbent.
Both measures are computed similarly to the previous two.
The confidence scores and the class support counts are compared by relational operators and the result is stored as binary matrices.
We sum their entries, normalise and compute the difference as before.
\\
The idea of the class support is to identify the hypothesis where the associated class label appears most often during the search.
Therefore, most transformed variations fall into one class cluster within the representation space. 
We argue that this might be evidence for the correctness of the prediction.
Also, an equivariant model would maximise this measure and assign all samples to one class. 

\subsection{Continuation of \cref{sec:siscore}}
\label{sec:siscore_cont}

\begin{table*}
  \centering
  \begin{tabular}{p{\columnwidth}lll}
    \toprule
    \centering Test Set Samples & ViT & acc@1 & acc@5 \\ \\
    \midrule
    \multirow{5}{*}{\begin{minipage}{0.5\columnwidth}\includegraphics[width=2.\columnwidth, height=.5\columnwidth]{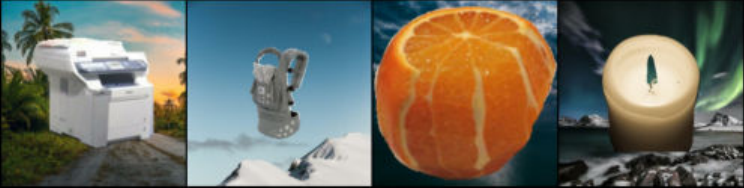}\end{minipage}} & vanilla & 54.9 & 77.5 \\
    & Rotation + Scale Augm. & 45.3 \textcolor{red}{(-9.6)} & 67.9 \textcolor{red}{(-9.6)} \\
    & \cite{Chefer2022} & 60.2 \textcolor{Gain}{(+5.3)} & 80.6 \textcolor{Gain}{(+3.1)} \\
    & \cite{Chefer2022}+AS & 61.0 \textcolor{Gain}{(+6.1)} & 81.4 \textcolor{Gain}{(+3.9)} \\
    & \gls*{its} & 55.1 \textcolor{Gain}{(+0.2)} & 77.7 \textcolor{Gain}{(+0.2)} \\
    & & & 72.3 \textcolor{red}{(-5.2)} \\
    \midrule
    \multirow{4}{*}{\begin{minipage}{0.5\columnwidth}\includegraphics[width=2.\columnwidth, height=.5\columnwidth]{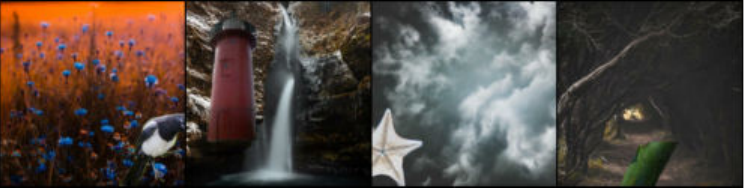}\end{minipage}} & vanilla & 33.3 & 52.2 \\
    & \cite{Chefer2022} & 38.6 \textcolor{Gain}{(+5.3)} & 57.0 \textcolor{Gain}{(+4.8)} \\
    & \cite{Chefer2022}+AS & 38.6 \textcolor{Gain}{(+5.3)} & 57.8 \textcolor{Gain}{(+5.6)} \\
    & \gls*{its} & 29.7 \textcolor{red}{(-3.6)} & 45.0 \textcolor{red}{(-7.2)} \\
    & & & 43.6 \textcolor{red}{(-8.6)} \\
    \midrule
    \bottomrule
  \end{tabular}
  \caption{Continuation of \cref{tab:siscore} with SI-Size and SI-Location.}
  \label{tab:siscore_cont}
\end{table*}

\begin{figure*}
    \centering
    \includegraphics[width=\textwidth]{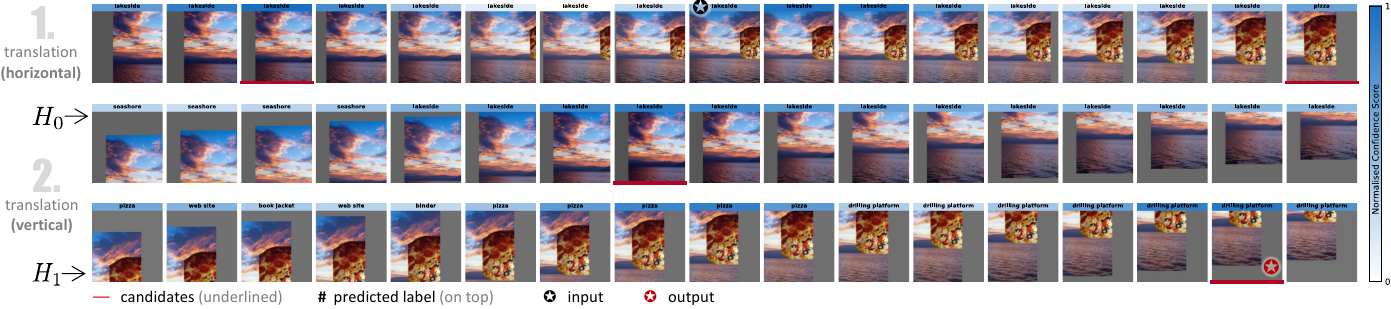}
    \caption{SI-Location sample showing a pizza (ground-truth label) piece on a lakeside background.
    In the first layer, $H_0$ classified the image as a lakeside and $H_1$ as a pizza.
    In the second layer, $H_0$ got refined but $H_1$ changed its hypothesis to now being a drilling platform.}
    \label{fig:focus-example1}
\end{figure*}

\cref{tab:siscore_cont} shows the continuation of our experimental results presented in \cref{tab:siscore}.
On SI-Size and SI-Location \gls*{its} is not able to push the performance of the baseline \gls*{vit}.
We found no improvement on both test sets.
One driving factor is the ability of \gls*{its} to focus on different parts of the image \emph{if} translation is used in at least one of the layers.
This "issue" is reinforced when another layer can scale the signal.
This creates a magnifying glass effect on objects, that the backbone networks predicts are the target objects.
As ImageNet classes are diverse, features might appear in the background leading the network to focus here.
In \cref{fig:focus-example1} we provide an example of a focus failure.

\newcommand{\STAB}[1]{\begin{tabular}{@{}c@{}}#1\end{tabular}}
\begin{table}
  \centering
  \begin{tabular}{llll}
    \toprule
    \centering & model & acc@1 & acc@5 \\ \\
    \midrule
    \multirow{4}{*}{\STAB{\rotatebox[origin=c]{90}{SI-Rotation}}} & EfficientNet & 25.1 & 43.0 \\
    & EfficientNet + ITS & 34.4 \textcolor{Gain}{(+9.3)} & 47.9 \textcolor{Gain}{(+4.9)} \\ 
    & ConvNeXt & 42.4 & 65.3 \\
    & ConvNeXt + ITS & 54.7 \textcolor{Gain}{(+12.3)} & 70.0 \textcolor{Gain}{(+4.7)} \\ 
    \midrule
    \multirow{4}{*}{\STAB{\rotatebox[origin=c]{90}{SI-Size}}} & EfficientNet & 35.7 & 57.3 \\
    & EfficientNet + ITS & 36.9 \textcolor{Gain}{(+1.2)} & 50.5 \textcolor{red}{(-6.8)} \\ 
    & ConvNeXt & 53.2 & 75.8 \\
    & ConvNeXt + ITS & 56.4 \textcolor{Gain}{(+3.2)} & 68.3 \textcolor{red}{(-7.5)} \\ 
    \midrule
    \multirow{4}{*}{\STAB{\rotatebox[origin=c]{90}{SI-Location}}} & EfficientNet & 15.7 & 30.3 \\
    & EfficientNet + ITS & 11.4 \textcolor{red}{(-4.3)} & 17.9 \textcolor{red}{(-12.4)} \\ 
    & ConvNeXt & 32.1 & 58.6 \\
    & ConvNeXt + ITS & 31.6 \textcolor{red}{(-0.5)} & 51.5 \textcolor{red}{(-7.1)} \\ 
    \bottomrule
  \end{tabular}
  \caption{Continuation of the SiScore experiment with an EfficientNet-B0 \cite{efficientnet} and a ConvNeXt-tiny \cite{convnext} backbone.}
  \label{tab:siscore_backbones}
\end{table}

Furthermore, we tested other popular backbone architectures, covering a EfficientNet \cite{efficientnet} and a ConvNeXt \cite{convnext}.
Both models were pretrained on ImageNet-1K.
The results are shown in \cref{tab:siscore_backbones}.

\section{Theoretical Specifications}
\label{sec:theory}

\subsection{Transformation Decomposition}
\label{sec:decomposition}

Group representations of $\operatorname{Aff}_2(\mathbb{R})$ are defined as $3 \times 3$ matrices with real entries \cite{hall2015lie}.
These matrices can be decomposed into a centre-rotation, isotropic scaling, reflection, shearing and translation matrix of the same size.
We use these basic (or atomic) transformation sub-groups to define the orbits of each search level.
\\
Centre-rotations and isotropic scaling can be represented by a single parameter.
Therefore, these sub-groups can be generated by a single generator.
For the remaining sub-groups, two generators are required.
Reflection is either done about the horizontal or vertical centre axis.
Shearing and translation are further decomposed as follows:
Either of these transformations is performed along and against the direction of some vector positions at the centre.



\subsection{Convolution over Finite Orbits}
With this decomposition, convolution over the orbit of any such sub-group can be defined.
This is essential for our group-induced confidence measure in \cref{eq:group_confidence}.

\begin{lemma}[Convolution over Finite Orbits] \label{lem:orbit}
    Convolving some function $f$ with the orbit, $\operatorname{orb}_G(\mathbf{x})$, can be expressed by
    \[
        \left(\operatorname{orb}_G(\mathbf{x})[n] \star f \right)(g) = \sum_{m} \operatorname{orb}_G(\mathbf{x})[m] f((\hat{g}^n)^{-1}\hat{g}^m).
    \]
\end{lemma}

The proof can be found in \cref{proof:orbit}.

\subsection{Markov Interpretation}
\label{sec:markov}

Each branching candidate $g^{-1}$ in the search process is a result of \cref{eq:inverse_approx}.
All of the aforementioned measures (see \cref{eq:energy,eq:epistemic,eq:group_confidence}) are stochastic when simulating Bayesian inference (as we do using Monte-Carlo dropout).
The evaluations are therefore based on a random variable, which allows us to formulate the process of traversing the search tree as a stochastic process.
\\
More specifically, the search describes a Markov process.
The Markov property of a stochastic process guarantees that the probability of a future state depends only on the current state independently of its past behaviour.
This implies that the branching candidate in each level only depends on the previous candidate.
This is true, as every roll-out depends only on the prior candidate in the first place.
These one-step dynamics simplify the system.
In \cref{proof:stochastic_process} we provide a sketch of the proof.


\section{Proofs}

\subsection{Proof: \cref{lem:orbit}}
\label{proof:orbit}

\begin{proof}
Given some kernel $K$ over the orbit of some finite group, $G$. Let $\operatorname{orb}_G(\mathbf{x})$ be the orbit of $\mathbf{x}$ under the action of $G$.
Convolving $\operatorname{orb}_G$ with $K$ is per definition given by
\begin{equation}
    \left(\operatorname{orb}_G(\mathbf{x})[g] \star K \right)(g) := \int_G \operatorname{orb}_G(\mathbf{x})[h] K(g^{-1}h) d h.
\end{equation}
As $G$ is a finite group, we can simplify the integration by the Riemann summation over $G$,
\begin{equation} \label{eq:orbit_conv_sum}
    \left(\operatorname{orb}_G(\mathbf{x})[g] \star K \right)(g) = \sum_{h \in G} \operatorname{orb}_G(\mathbf{x})[h] K(g^{-1}h).
\end{equation}
This can be rewritten using generators, such that
\begin{equation} \label{eq:conv_orbit_gen}
    \left(\operatorname{orb}_G(\mathbf{x})[n] \star K \right)(g) = \sum_{m} \operatorname{orb}_G(\mathbf{x})[m] K((\hat{g}^n)^{-1}\hat{g}^m).
\end{equation}
It is to be shown that $\operatorname{orb}_G(\mathbf{x})$ can be evaluated by $n \in \mathbb{N}$ instead of $g$.
This needs to be justified for the single and dual generator case.

\paragraph{Single Generator}
For cyclic groups, where one generator suffices, every $g \in G$ can be substituted by the generator raised to some power. 
The orbit map from \cref{def:orbit} can be rewritten as
\begin{align}
    \operatorname{orb}_G(\mathbf{x}) &=  \{ g\mathbf{x} \mid g \in G \} \nonumber \\
    &= \{ \hat{g}^n\mathbf{x} \mid n \in [0, 1, \ldots, |G|-1] \}.
\end{align}
where $\hat{g}^n$ denotes the generator raised to power $n$.
This isomorphism between $g$ and $n$ also simplifies the evaluation
\begin{equation}
    \operatorname{orb}_G(\mathbf{x})[g] = \operatorname{orb}_G(\mathbf{x})[n].
\end{equation}

\paragraph{Dual Generator}
The second case assumes, that $G$ is non-cyclic and requires two generators, $\hat{g}$ and $\hat{h}$.
Let $\hat{g}$ be the generator for the positive direction along some axis and $\hat{h}$ the generator for the negative direction.
Hence, elements in $G$ are either generated by $\hat{g}$ or $\hat{h}$.
This allows us to define the following generating function
\begin{equation}
    \phi(n) = 
    \begin{cases}
        \hat{g}^n,& \text{if } n\geq 0\\
        \hat{h}^n,              & \text{otherwise.}
    \end{cases}
\end{equation}
This generalizes the orbit definition to
\begin{align}
    \operatorname{orb}_G(\mathbf{x}) &=  \{ g\mathbf{x} \mid g \in G \} \nonumber \\
    &= \{ \phi(n)\mathbf{x} \mid n \in [0, 1, \ldots, |G|-1] \}.
\end{align}
\end{proof}

To further extend this idea, we can also let any function $f(\mathbf{x}): \operatorname{orb}_G(\mathbf{x}) \to \mathbb{R}$ act on the orbit.
We can define $f$ to be an Energy function, $\bar{E}_{\mathbf{\theta}}(\mathbf{x})$.
This forms the foundation for our group-induced confidence measure, $\rho_G(\mathbf{x}, g)$, in \cref{eq:group_confidence}.

\subsection{Proof: \cref{thm:group_confidence}}
\label{proof:group_confidence}

\begin{proof}
Let $E_{\mathbf{\theta}}(\mathbf{x})$ be the energy of a $\theta$-parameterised energy model given a point $\mathbf{x}$ as input as defined in \cref{eq:energy}.
Let $\bar{E}_{\mathbf{\theta}}(\mathbf{x})$ be the Monte-Carlo estimate of the energy using \cref{eq:mcd_expectation}.
As shown in \cref{lem:orbit} we can convolve this energy function on the orbit with a kernel, $K$.
To remove visual clutter, we substitute
\begin{equation}
    \phi(\mathbf{x}, g) := \big( \bar{E}_{\mathbf{\theta}} \star K \big)\big(g\mathbf{x}\big).
\end{equation}
We are interested in the second derivative of $\phi(\mathbf{x}, g)$ w.r.t. some $g \in G$.
We use the second-order Taylor polynomial and assume sufficiently small remainders.
Hence, allowing us to approximate the derivative of $\phi$ by infinitesimal small steps,
\begin{align}
    \phi(\mathbf{x}, g+h) &\approx
    \phi(\mathbf{x}, g)
    + \phi^\prime(\mathbf{x}, g) h
    + \frac{1}{2} \phi^{\prime\prime}(\mathbf{x}, g) h^2 \\
    \phi(\mathbf{x}, g-h) &\approx
    \phi(\mathbf{x}, g)
    - \phi^\prime(\mathbf{x}, g) h
    + \frac{1}{2} \phi^{\prime\prime}(\mathbf{x}, g) h^2.
\end{align}
Adding both together and rearranging for $\phi^{\prime\prime}$ gives
\begin{align} \label{eq:taylor}
    &\phi(\mathbf{x}, g+h) + \phi(\mathbf{x}, g-h) \approx
    2\phi(\mathbf{x}, g) 
    + \phi^{\prime\prime}(\mathbf{x}, g) h^2 \nonumber \\
    &\phi^{\prime\prime}(\mathbf{x}, g) \approx 
    \frac{1}{h^2} \bigg( 
    \phi(\mathbf{x}, g+h) 
    + \phi(\mathbf{x}, g-h)
    - 2\phi(\mathbf{x}, g)
    \bigg).
\end{align}
This formulation uses incrementally small steps $g+h$ on the orbit of $G$ and $\mathbf{x}$, which we have to define next.
\\
Given a Matrix Lie Sub-Group, $H$ and a finite sub-group, $G \leq H$.
We can approximate $H$ by increasing the cardinality of $G$, such that in the limit $G=H$.
Let $\langle g \rangle$ be the generator of $G$, which forms an atomic element of $G$.
Then, every $g \in G$ can be expressed by $g = \hat{g}^k$ for some $k$. 
Using composition, a step along the orbit starting at $g$ can be expressed by $(\hat{g} \circ g) \mathbf{x}$.
Similarly, a step in the opposite direction can be performed by $(\hat{g}^{-1} \circ g) \mathbf{x}$.
We substitute $g\mathbf{x}+h$ by $(\hat{g} \circ g) \mathbf{x}$ in \cref{eq:taylor}.
\\
Let $\mathbf{T}_g$ be the matrix representation of $g$.
The step length along the orbit is defined by $\| \mathbf{T}_{\hat{g}} - \mathbf{T}_{\hat{g}^2} \|$.
These two observations lead to the following reformulation of \cref{eq:taylor},
\begin{align}
    &\phi^{\prime\prime}(\mathbf{x}, g) \approx \frac{1}{\| \mathbf{T}_{\hat{g}} - \mathbf{T}_{\hat{g}^2} \|^2} \\
    &\bigg( \phi(\mathbf{x}, \hat{g} \circ g) + \phi(\mathbf{x}, \hat{g}^{-1} \circ g) - 2\phi(\mathbf{x}, g)\bigg) \nonumber.
\end{align}
Remember that $G$ is a finite sub-group of some Matrix Lie Group, $H$.
In the limit, however, when the cardinality of $G$ approaches the one of $H$, the approximation error will shrink to zero.
Therefore, we consider
\begin{align}
    &\phi^{\prime\prime}(\mathbf{x}, g) = \lim_{|G| \to \infty} \frac{1}{\| \mathbf{T}_{\hat{g}} - \mathbf{T}_{\hat{g}^2} \|^2} \\
    &\bigg( \phi(\mathbf{x}, \hat{g} \circ g) + \phi(\mathbf{x}, \hat{g}^{-1} \circ g) - 2\phi(\mathbf{x}, g)\bigg) \nonumber.
\end{align}
This concludes the proof.
\end{proof}

\subsection{Proof: \cref{thm:group_elimination}}
\label{proof:group_elimination}

\begin{proof}
Given a finite sequences $s = g_0 \circ g_1 \circ \ldots \circ g_m$ of group elements and a sub-sequence $s_{0:k-1} = g_0 \circ g_1 \circ \ldots \circ g_{k-1}$ with $k \leq m \in \mathbb{N}$.
It is to be shown that additional action by $g_k^{-1}$ on $s$ eliminates $g_k$ from $s$ (this sequence is coined $s_{\setminus g_k}$).
We prove the statement $g_k^{-1}s = s_{\setminus g_k}$ by successive refinement and ignore the trivial case where $k = m$, $k=0$ or $m=0$.
We have
\begin{equation}
    g_k^{-1}s = g_k^{-1} \circ (g_0 \circ g_1 \circ \ldots \circ g_m).
\end{equation}
Due to the associativity axiom of any group, the parentheses can be ignored.
Furthermore, $s$ can be substituted by $s_{0:k-1} \circ s_{k:m}$, such that
\begin{equation}
    g_k^{-1}s = g_k^{-1} \circ \underbrace{g_0 \circ \ldots \circ g_{k-1}}_{s_{0:k-1}} \circ \underbrace{g_k \circ \ldots \circ g_m}_{s_{k:m}}.
\end{equation}
If elements $s_{0:k-1}$ are non-trivial, then commutativity is required to move $g_k^{-1}$ in front of $g_k$ to cancel them out.
Therefore, we need to assume that $s_{0:k-1}$ and $g_k^{-1}$ are elements of an Abelian group.
Then,
\begin{align}
    g_k^{-1}s &= s_{0:k-1} \circ g_k^{-1} \circ g_k \circ \ldots \circ g_m, \\
    g_k^{-1}s &= s_{0:k-1} \circ \ldots \circ g_m, \\
    g_k^{-1}s &= s_{\setminus g_k},
\end{align}
which concludes the proof.
\end{proof}

\subsection{Proof: Markov Process}
\label{proof:stochastic_process}

We begin by formally defining a stochastic and a Markov process.

\begin{definition}[Stochastic Process \cite{Borodin2017}]
A stochastic process is a family of random variables $\{ X_t : t \in \mathcal{T} \}$, where the index $t$ belongs to some arbitrary parameter set, $\mathcal{T}$.
Each of the random variables can take the same set of possible values, which are called the \emph{state space}.
\end{definition}

\begin{definition}[Markov Process \cite{Borodin2017}]
Given a stochastic process, $\{ X_t : t \in \mathcal{T} \}$.
We say $\{ X_t : t \in \mathcal{T} \}$ has the Markov property if for all $t$ and all states $\{ x_i : i \in [0, n] \}$, we have
\begin{align}
    \mathbb{P}\left(X_{n+1}=x_{n+1} \mid \{X_i=x_i : i \in [0, n]\}\right) \nonumber \\
    =\mathbb{P}\left(X_{n+1}=x_{n+1} \mid X_n=x_n\right).
\end{align}
\end{definition}

When traversing the search tree, a composition of transformation matrices is formed, $\mathbf{T}_t \circ \mathbf{T}_{t+1} \circ \ldots \mathbf{T}_{T}$.
These matrices form the state space.
At each state, the group-induced confidence measure, $\rho_G(\mathbf{x}, g)$, is used to determine the next most probable action.
To reduce visual clutter, we define $g_t := g \in G_t$, i.e., some group element from the group at level $t$.
Given a finite set of actions, we use the Boltzmann distribution (softmax) to formulate state-action probabilities,
\begin{align}
    \mathbb{P}(\mathbf{T}_{t} \mid \hat{\mathbf{x}}) 
    &= \frac{\exp \rho_{G_t}(\hat{\mathbf{x}}, g_t)}{\sum_{g_t} \exp \rho_{G_t}(\hat{\mathbf{x}}, g_t)} \\
    &= \frac{\exp -\frac{\partial^2}{\partial g_t^2} \big( \bar{E}_{\mathbf{\theta}}(g_t\hat{\mathbf{x}}) \star K \big)}{\sum_{g_t} \exp -\frac{\partial^2}{\partial g_t^2} \big( \bar{E}_{\mathbf{\theta}}(g_t\hat{\mathbf{x}}) \star K \big)}, \nonumber
\end{align}
where $\hat{\mathbf{x}}$ represents $\mathbf{x}$ after all previous group elements on the trajectory acted on it, i.e., $\hat{\mathbf{x}} := (g_{t-1} \circ g_{t-2} \circ \ldots \circ g_0) \mathbf{x}$.
The mapping from $\mathbf{x}$ to $\hat{\mathbf{x}}$ is surjective.
Trivially, the identity can be injected at all positions without changing $\hat{\mathbf{x}}$.
The same is true for (non-trivial) equivariance classes.
As long as $\hat{\mathbf{x}}$ stays the same, the conditioning $\mathbb{P}(\mathbf{T}_{g_t} \mid \hat{\mathbf{x}})$ will not be affected by the choice of transformation sequence.
Due to this finding, we have
\begin{align}
    \mathbb{P}(\mathbf{T}_{t} \mid \hat{\mathbf{x}}) = \mathbb{P}(\mathbf{T}_{t} \mid \mathbf{T}_{t-1}).
\end{align}
This conditioning fulfils the definition of a Markov property.
\\
Next, we complete this proof by showing that we have a stochastic process.
$\bar{E}_{\mathbf{\theta}}(\cdot)$ is the expected energy obtained using approximate Bayesian inference,
\begin{align}
    \mathbb{E}_{\mathbf{\Lambda}}[E_{\mathbf{\theta}}(\mathbf{x})] := \bar{E}_{\mathbf{\theta}}(\mathbf{x}) = \int z \mathbb{P}_{\mathbf{\theta}}(z \mid \mathbf{x}, \mathbf{\Lambda}) d z,
\end{align}
where $z$ is a latent scalar for the total energy.
As defined in the main corpus of this work, we use Monte-Carlo Dropout to simulate multiple stochastic forward passes, where $\Lambda := \operatorname{diag}(\boldsymbol{\lambda}) \sim \text{Bernoulli}(\lambda \in [0, 1])$.
So, the measure itself is a random variable sampled from a distribution defined by the neural network.
We use kernel density estimation with Gaussian kernels to estimate the energy probabilities, $\mathbb{P}_{\mathbf{\theta}}(z \mid \mathbf{x}, \mathbf{\Lambda})$.
It is this stochastic measure which renders the process a stochastic process.
The transition matrix is formed by $\mathbb{P}(\mathbf{T}_{t} \mid \mathbf{T}_{t-1})$ for all $t$.
We are only interested in the forward pass, i.e., traversing the tree from root to leaf.
As, however, all group actions are reversible per definition (see \cref{def:axioms}), the backward pass is well-defined, too.

\end{document}